\def\year{2021}\relax
\documentclass[letterpaper]{article} 
\usepackage{aaai21}  
\usepackage{times}  
\usepackage{helvet} 
\usepackage{courier}  
\usepackage[hyphens]{url}  
\usepackage{graphicx} 
\usepackage{natbib}  
\usepackage{caption} 
\usepackage{booktabs}
\usepackage{multirow}
\usepackage{multicol}

\usepackage{amsmath}
\usepackage{amsfonts}
\usepackage{amssymb}
\usepackage{amsthm}

\usepackage{xcolor}

\theoremstyle{definition}

\newtheorem{theorem}{Theorem}[section]

\newcommand{\newcite}[1]{\citeauthor{#1}~(\citeyear{#1})}
\newcommand{\genre}{type}
\newcommand{\genres}{types}
\newcommand{\Genre}{Type}
\newcommand{\type}{value}
\newcommand{\types}{values}

\title{Multi-\genre\ Disentanglement without Adversarial Training}

\author{Lei Sha,\ \  Thomas Lukasiewicz\\}
\affiliations{
Department of Computer Science, University of Oxford, UK \\
  \texttt{firstname.lastname
 @cs.ox.ac.uk}
}
\date{}

\begin{document}
\maketitle
\begin{abstract}
Controlling the style  of natural language by disentangling the latent space is an important step towards interpretable machine learning. After the latent space is disentangled, the style of a sentence can be transformed by tuning the style representation without affecting other features of the sentence.
Previous works usually use adversarial training to guarantee that disentangled vectors do not affect each other. 
However, adversarial methods are difficult to train. Especially when there are multiple features (e.g., sentiment, or tense, which we call \textit{style \genres} in this paper), each feature requires a separate discriminator  for extracting a disentangled style vector corresponding to that feature. 
   In this paper\footnote{The code for this paper can be found at \url{https://sites.google.com/site/codeforpaper/home}}
, we propose a unified distribution-controlling method, which provides each specific \textit{style \type} (the value of style \genres, e.g., positive sentiment, or past tense) with a unique representation. This method contributes a solid theoretical basis to avoid adversarial training in multi-\genre\ disentanglement. We also propose multiple loss functions to achieve a style-content disentanglement as well as a disentanglement among multiple style \genres. 
In addition, we observe that if two different style \genres\ always have some specific style \types\    that occur together in the dataset, they will affect each other when transferring the style \types. We call this phenomenon \textit{training bias}, and we propose a loss function to alleviate such training bias while disentangling multiple \genres. 
We conduct experiments on two datasets (Yelp service reviews and Amazon product reviews) to evaluate the style-disentangling effect and the unsupervised style-transfer performance on two style \genres: sentiment and tense. 
The experimental results show the effectiveness of our model.
 
\end{abstract}  

\section{Introduction}
Changing some specific features (such as sentiment, tense, or human face pose) of a given text or image is very important in  many applications. These features are usually embedded in the weights of  ``black-box'' deep neural networks. Hence, disentangling the latent space of the neural network becomes a valuable step of such tasks.

In the early years, researchers tried to use some disentangled latent variables to control latent features of images and text. For example, \newcite{chen2016infogan} use scalar latent variables to control the writing style for handwritten digits as well as the pose of human face 3D-rendered images by maximizing the mutual information between the latent variable and the generator.  
Furthermore, there is previous research that tends to completely separate the latent space into disentangled components. For example, \newcite{john-etal-2019-disentangled} use multiple adversarial optimizers to decrease the dependency between the latent vectors of style and content.


A severe limitation in previous works is adversarial training, which is always difficult to train and usually requires many resources. Especially when we are extracting multiple style vectors and each of them represents a specific feature, then for each kind of feature, there needs to be a discriminator,  according to \newcite{john-etal-2019-disentangled}. Therefore, a generator with so many discriminators will make the training process extremely complicated. 

In this paper, we distinguish the concepts of style type and value: (1) a \textit{style \genre} is a  style class that represents a specific feature of text or an image, e.g., sentiment, tense, or face direction; and (2) a \textit{style \type} is one of the different values within a style \genre, e.g., sentiment (positive/negative), or tense~(past/now/future).   We propose a unified distribution-controlling method that gives a unique representation to each style \type\ in each style \genre. We assume that the representations of each style \type\ are sampled from separate Gaussian distributions. 
Then, we generate multiple style vectors (for different style \genres) and a content vector using the input text,  and force the  style vectors to be close to the ground-truth style-\type\ distributions. To ensure that the semantic information is not lost, we sample a style vector from the ground-truth style-\type\ distribution for each style \genre, and combine all the style vectors  and the  content vector into one vector. Then, we force the combined vector to reconstruct the original sentence. 
To avoid adversarial training, we propose a loss function for style-content disentanglement to make it more efficient, which is applicable to both vanilla and variational autoencoders. We further propose loss functions for the disentanglement among multiple style \genres. In addition, we point out a severe problem in multi-\genre\ disentanglement,  which is called \textit{training bias}. We prove mathematically that our multi-\genre\ disentanglement loss function can alleviate the training bias problem.

We conducted experiments on two datasets: Yelp Service Reviews
~\cite{Shen2017Style} and Amazon Product Reviews~\cite{Fu2018Style}. The experimental results show that our method can provide a comparable disentanglement effect and even better style-transfer effect without the help of adversarial training.
The experimental results also show our method's effectiveness for alleviating the training bias.

The contributions of this paper are briefly as follows:
\begin{itemize}
\item We propose a unified distribution-controlling method for disentanglement, which provides unique representations for each style \type\ in each style \genre\ and provides a natural advantage for multi-\genre\ disentanglement.
\item We propose loss functions to disentangle style and content without adversarial training. Our method is  applicable even in the situation where one \genre\ contains multiple  style \types, both   in  vanilla  and variational autoencoders.
\item We propose loss functions for disentangling multiple style \genres, which can also alleviate the bias caused by multi-\genre\ training data. Based on a solid theory,  the  effectiveness of these loss functions
is also shown in experiments.
\end{itemize}

The rest of this paper is organized as follows.
Section~\ref{sec:approach} introduces the details of our approach and the losses designed for style-content 
and multi-\genre\ disentanglement. 
We then conduct experiments in  Section~\ref{sec:exp}, discuss related work in Section~\ref{sec:rel},  and finally conclude our work in Section~\ref{sec:con}.

\section{Approach}\label{sec:approach}

 In this section, we first make an assumption about style vectors, called \textit{unified distribution assumption}. On top of this, we then build our model architecture in Fig.~\ref{fig:arch}. We further propose two loss functions for style-content disentanglement (which encourage that the style vectors and the content vector do not affect each other) and multi-\genre\ disentanglement (which reduces the effect among the style vectors).

\subsection{Unified Distribution-Controlling Method}

Intuitively, for the sentences belonging to one specific style \type, the style vector generated by them should have the same representation. Since this is a very strong condition, we use a relaxed form of this requirement as \textit{unified distribution assumption}, in which we require that all the vectors that belong to one specific style \type\ follow a unified  distribution, the parameters of which can also be updated during training. We use a Gaussian distribution with different mean and variance for each style \type. Then, we require the disentangled vectors to satisfy the following requirements.
\begin{itemize}
\item Disentangled vectors that belong to the same style \type\ should obey the same Gaussian distribution. 
\item The Gaussian distributions corresponding to any two different style \types\ should be independent from each other.
\end{itemize}
The control under the {unified distribution assumption} is as follows.  When we are conducting style transfer, we only need to sample a new style vector from the target style \type\  and replace the original one, then the corresponding style of the generated sentence will also be transferred.

\subsection{Model}
\begin{figure}
    \centering
    \includegraphics[width=\linewidth]{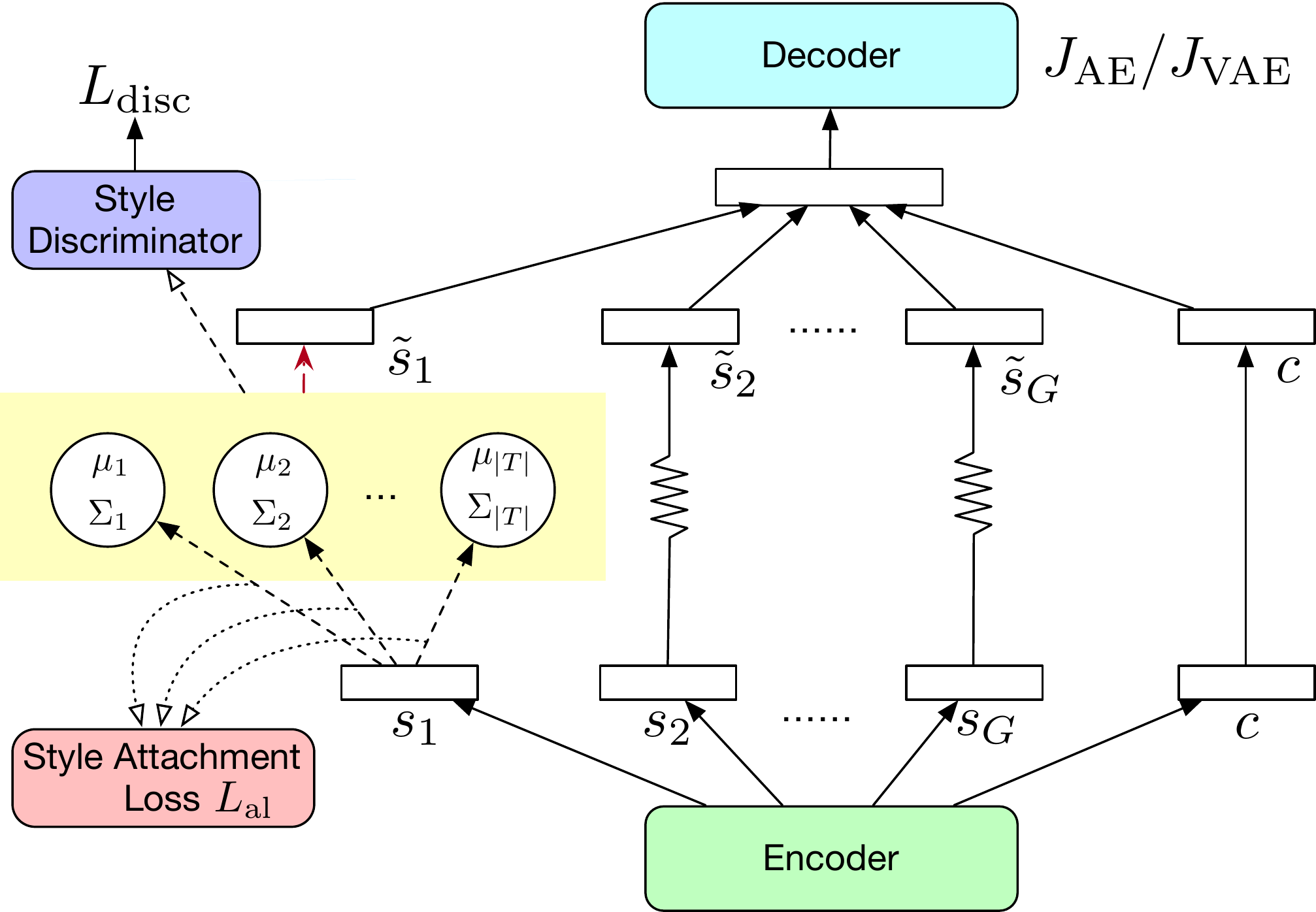}
\smallskip 
    \caption{Architecture of our method: the circles labeled with  $\mu_{i}$ and $\Sigma_{i}$ represent  the style-\type\ distributions, while the zig-zag arrows are the omission of the sampling process from the correct style \type.}
    \label{fig:arch}
\end{figure}

We now describe our model in detail. As shown in Fig.~\ref{fig:arch}, the basic architecture of our method is an autoencoder. We first encode the input sentence and generate several disentangled style vectors and a content vector from the encoded sentence. The style vector is required to be close to the correct unified style distribution.
Ideally, after the sentence is perfectly disentangled into style  and content vector, if the style is replaced by the unified representation of the same style \type, the original sentence can be correctly reconstructed by the new style vector and the original content vector.
 Therefore, in the final step, we sample a vector from the correct unified style distribution to replace the original one and reconstruct the original sentence.

\subsubsection{Autoencoder.}
In disentanglement approaches, the best way to prevent the loss of information is using autoencoders, which encode the input sentence to a latent space and then reconstruct the original sentence from this space. In our method, we use two kinds of autoencoders:  a vanilla and a variational autoencoder~\cite{kingma2014auto}.
Each training example is composed of a sentence $x$ and a bunch of  style \types\ corresponds to different style \genres: $t_{1\star},\ldots,t_{G\star}$\footnote{We use $t_{ij}$ to represent the $j$-th \type\ of the $i$-th style \genre, and ``$\star$'' means that we have the same action for each style \genre~(when it replaces $i$) or \type~(when it replaces $j$). }  ($G$ is the number of style \genres). Given the input token sequence $x=\{x_1, x_2,\ldots, x_n\}$, we use an  LSTM-based~\cite{hochreiter1997long} vanilla and a variational autoencoder to build the reconstruction loss:
\begin{equation}
\begin{small}
\begin{aligned}
  J_\text{AE} =& \mbox{$-\sum_t\log P(x_t|h, x_1, x_2,\ldots, x_{t-1})$}\\
  J_\text{VAE} =&-\int q_E(h|x)\log [P(x|h)]dh \\
  &+ \lambda_\text{KL}\text{KL}(q_E(h|x)||P(h)),\\
\end{aligned}
\end{small}
\end{equation}
where $h$ is the latent vector, $P(x|h)$ is  the decoder part, $P(h)$ is the prior distribution of $h$ (usually,  $\mathcal N(0,1)$), and $\lambda_\text{KL}$ is a hyperparameter.
The latent vector is then split into $G$  style vectors  $s_1,\ldots,s_G$ of equal length and a content vector $c$.

%

\subsubsection{Style Attachment Loss on Latent Style Space.}
For any style \genre, we need to make the style vector $s_\star$ ``appear like'' sampled from the correct style \type's  unified distribution. 
For simplicity, we omit the subscripts and use $s$ for style vector,  and $t$ for style \type\ in this section. So, we need to maximize the probability of the style vector $s$ belonging to style \type\ $t$, denoted $P(t|s)$.

 In Fig.~\ref{fig:arch}, the Gaussian distributions of the style \types\ have parameters $\mu_{\star j}$, $\Sigma_{\star j}$, $j\,{\in}\, \{1,\ldots,|T|\}$ ($T$ represents the set of style \types, $|T|$ represents the number of style \types). Then, the probability density function (PDF) of the $j$-th style \type\ $T_{\star j}$ is as follows:
 \begin{equation}\label{eq:gauss}
\begin{small}
\begin{aligned}
   p_\text{Nor}(s|T_{\star j})=\frac{\exp\Big(-\frac{1}{2}(s-\mu_{\star j})\Sigma_{\star j}^{-1}(s-\mu_{\star j})\Big)}{\sqrt{(2\pi)^d \det(\Sigma_{\star j})}},
   \end{aligned}
   \end{small}
 \end{equation}
 where $d$ means the dimension of the style vector, and ``Nor'' is short for ``Normal distribution''. Then, we use Bayes' theorem to  calculate this probability as  shown in Eq.~\ref{eq:asmsfm}:
\begin{equation}\label{eq:asmsfm}
  P(t|s) = \frac{p_\text{Nor}(s|t)p(t)}{p(s)} = \frac{p_\text{Nor}(s|t)p(t)}{\sum_{t'\in T}  (p_\text{Nor}(s|t')p(t'))},
\end{equation}
where $p(t)$ is the prior distribution of the style \types, which is decided by the dataset.

Therefore, we define the style attachment loss as the negative log-likelihood (NLL) loss according to the labeled style \type:
\begin{equation}
 \mbox{$ L_\text{al} = -\frac{1}{|D|}\sum_{m=1}^{|D|}\log P(t^{(m)}|s^{(m)})$,}
\end{equation}
where $|D|$ denotes the size of the training set, and  $t^{(m)}$ represents  the label of the $m$-th case in the training set.
 
%

\subsubsection{Style Classification Loss.}
We need to make each unified distribution really map to the corresponding style \type, so we sample vectors from each style \type\ distribution and force them to be classified to that style \type. 

We still use the Gaussian distribution to calculate the classification loss.
We first sample $M$ vectors  from the distributions $\mathcal N(\mu_{\star j},\Sigma_{\star j})$, denoted $\tilde s^{(m)}_{\star j}$ (which is  the $m$-th sample from the $j$-th style \type\ distribution). Then, we calculate the probability of $\tilde s^{(m)}_{\star j}$ for the style \type\ $T_{\star j}$~as~in~Eq.~\ref{eq:pc}:
\begin{equation}\label{eq:pc}
  P_\text{c}(T_{\star j}|\tilde s^{(m)}_{\star j}) = \frac{p_\text{Nor}(\tilde s^{(m)}_{\star j}|T_{\star j})}{\sum_{t'\in T} p_\text{Nor}(\tilde s^{(m)}_{\star j}|t')}.
\end{equation}

Since the distribution's parameters $\mu_{\star j}$ and $\Sigma_{\star j}$  also need to be updated in the training phase, we use a reparameterization trick~\cite{devroye1996random,kingma2014auto} to make the sampling process differentiable: 
\begin{equation}
  \epsilon\sim\mathcal N(\textbf{O},\textbf{I}),\quad \tilde s^{(m)}_{\star j} = A_{\star j}\epsilon+\mu_{\star j},\\
\end{equation}
where $\textbf{O}$ is an all-zero matrix, $\textbf{I}$ is an identity matrix, and $A_{\star j}A_{\star j}^\top\,{=}\,\Sigma_{\star j}$. So, we take $A_{\star j}$, $j\,{\in}\, \{1,2,\ldots,|T|\}$, as the distribution parameters that need to be trained instead of $\Sigma_{\star j}$.
Then, we define the classification loss as an NLL~loss:
\begin{equation}
\mbox{$  L_\text{cl} = -\frac{1}{M|T|}\sum_{j=1}^{|T|}\sum_{m=1}^{M}\log P_\text{c}(T_{\star j}|\tilde s^{(m)}_{\star j})$.}
\end{equation}

\subsection{Style-Content Disentanglement}\label{sec:sc}
We need to guarantee that the content vector does not contain anything about the style. 
Since we need to disentangle the content vector with both the style vectors before and after the style \type\ sampling, we propose to minimize two pieces of mutual information: $I(c,t)$ between the content vector $c$ and the style labels $t$, and  $I(c,s)$ between the content vector~$c$ and the style vectors $s$. 
\paragraph{$\textbf{I(c,t)}$:}
To minimize the mutual information $I(c,t)$, we only need to minimize its upper bound, which can be stated as follows (a detailed proof is given in the extended paper):
\begin{equation}\label{eq:ict}
\begin{small}
\begin{aligned}
I(c,t)&=\mbox{$\mathbb{E}_x\Big[\int_c\sum_{t}p(c,t|x)\log\frac{p(c,t|x)}{p(c|x)p(t)}\Big]$}\\
&\le\mbox{$\mathbb{E}_x\Big[\sum_{t'}p(t')KL(p(c|t,x)||p(c|t',x))\Big]$,}\\
\end{aligned}
\end{small}
\end{equation}
where $p(t')$ is a constant in specific datasets,  $p(c|t,x)$ need to be modeled by another Gaussian distribution $\mathcal N_c(\mu'_t,\Sigma'_t)$, and then we force all the content vectors with label $t$ to obey this distribution.
To achieve that, we minimize the negative log-likelihood of the style labels in a batch.
\begin{equation}\label{eq:probnll}
\mbox{$L_\text{prob-nll}=-\frac{1}{M}\sum_{m=1}^M\log p_c(c^{(m)}|t^{(m)})$,}
\end{equation}
where $p_c(c^{(m)}|t^{(m)})$ is obtained from the Gaussian distribution $\mathcal N_c(\mu'_t,\Sigma'_t)$ in a  similar way  as $p_\text{Nor}(s|T_{\star j})$ in Eq.~\ref{eq:gauss}, and~$M$ represents  the batch size.

So, the loss function of style-content disentanglement is shown as follows.
\begin{equation}
\begin{aligned}
L_\text{sc}&= \mbox{$\mathbb{E}_x\Big[\sum_{t'}p(t')KL(p_c(c|t,x)||p_c(c|t',x))\Big]$}\\
     &+\lambda_\text{sc}L_\text{prob-nll}, \\
\end{aligned}
\end{equation}
where $\lambda_\text{sc}$ is a predefined hyperparameter, and  $p_c(c|t)$ is constrained by the loss in Eq.~\ref{eq:probnll}. The KL divergence item is tractable, because $p_c$ is a Gaussian distribution.



According to the final form of $L_\text{sc}$, our method is similar to the previously proposed variational fair autoencoder~\cite{louizos2015variational}. In their method, they propose a maximum mean discrepancy penalty as a regularizer to the model, which encourages the statistical moments of two classes to be the same. Differently from them, we build a prior distribution to each label and encourage these distributions to be the same, which is easy to apply to multi-class cases.  On the other hand, the method of compressing the input $x$ out of $c$~\cite{moyer2018invariant} is more eligible on variational autoencoders. In comparison, our derivation of $I(c,t)$ is based on our Gaussian assumption, which is eligible for both vanilla and variational autoencoder architectures.

\paragraph{$\textbf{I(c,s)}$:}
\label{sec:ics}
We prove (in the extended paper) that minimizing $I(c,s)$ is equal to minimizing $KL(p(c|x)||p(c))$ and $KL(p(s|x)||p(s))$, which is just the regularization term in variational autoencoders. So, we do not have any loss function for $I(c,s)$.

\subsection{Multi-\genre\ Disentanglement}\label{sec:mg}
There are always multiple style \genres\  occurring in a text or image, such as sentiment polarity and tense. We tend to solve two problems in this section: (1) make the style vectors $s_1,\ldots,s_G$ ($G$ is the number of style \genres) for different style \genres\ independent to each other, and (2) when the training set has labels of multiple \genres, there will be a \textit{training bias} that makes different \genres\ affect each other. We can write the training bias as $p(T_{i\star}|T_{j\star})>p(T_{i\star}), i\ne j$, where $T_{i\star}$ and $T_{j\star}$ stand for style \type\ labels in style \genres\ $i$ and $j$, respectively.\footnote{Ideally, it should be $p(T_{i\star}|T_{j\star})=p(T_{i\star}), i\ne j$.}  For example, if positive sentiment always occurs together with the past tense, then the positive sentiment style vector tends to carry information of the past tense.

Implicit disentanglement~\cite{higgins2017beta,chen2018isolating} uses unsupervised methods for scalar multi-\genre\ disentanglement, where each dimension of the latent vector encodes a specific feature (style \genre). 
Inspired by $\beta$-TC\-VAEs~\cite{chen2018isolating}, multi-\genre\ vector disentanglement can also be done by minimizing the \textit{total correlation} term:
\begin{equation}\label{eq:ltc}
\mbox{$KL(q(s_1,\ldots,s_G)||\prod_i q(s_i))$,}
\end{equation} 
where $G$ means the number of different \genres. 


In our unified distribution  settings, the $s_i$'s are generated by their own style-\type\ distribution instead of the input~$x$. 
So,  $q(s_1,s_2\ldots,s_G)$ can  be factorized as Eq.~\ref{eq:qsss} ($T_{1x},\ldots,$ $T_{Gx}$ are the corresponding style \types\ of $x$ in $G$ \genres):
\begin{equation}\label{eq:qsss}
\begin{aligned} 
\mbox{$q(s_1,s_2\ldots,s_G)$}& = \mbox{$\mathbb E_x[\prod_iq(s_i|T_{1x},\ldots,T_{Gx})]$}\\
&=\mbox{$\sum_x\Big[\frac{\prod_i^Gq(T_{1x},\ldots,T_{Gx}|s_i)q(s_i)}{p(x)^{G-1}}\Big]$.}\\
\end{aligned}
\end{equation}
The proof is shown in the extended paper. 
Usually, since $T_{ix}$ are potentially related, we have:
\begin{equation}
\begin{small}
\begin{aligned}
\mbox{$q(T_{1x},\ldots,T_{Gx}|s_i)=\prod_j^G q(T_{jx}|s_i,T_{kx(k=1\ldots G,k\ne j)})$.}
\end{aligned}
\end{small}
\end{equation}
Then, we have the following theorem (proved in the extended paper) to achieve multi-\genre\ disentanglement. Here, $\mathcal H(\cdot)$ denotes the entropy of a probability distribution.

\begin{theorem}\label{the:tts}{\it 
For random vector variables $s_1, \ldots, s_G$ and \types\ $t_1,\ldots, t_G$ sampled from G style \genres\footnote{For clarity, $t_i$ is a random variable, $T_{ix}$ is a constant label.}, if $\mathcal H(p(t_i|s_i))\,{=}\,0$, for all $i$, and  $\mathcal H(p(t_j|s_i))$ $=$ MAX,\footnote{We use ``=\,MAX" to represent ``reaches the maximum value".} for all $i,j$ with $j\ne i$, then for all $i,j$ with $j\ne i$, it holds that $p(s_j|t_i)=p(s_j)$ and $p(t_i|t_j,s_i)=p(t_i|s_i)$.
}\end{theorem}

To alleviate the training bias, we  also need to ensure that $p(t_i|t_j)=p(t_i)$ instead of just  $p(t_i|t_j,s_i)=p(t_i|s_i)$. So, we also need to make  $\mathcal H(p(t_j|t_i))$, for all $i,j$ with $j\ne i$,  reach the maximum value.

According to Theorem~\ref{the:tts}, when we guarantee that $\mathcal H(p(t_i|s_i))=0$, for all $i$, and $\mathcal H(p(t_j|s_i))=\text{MAX}$ and $\mathcal H(p(t_j|t_i))=\text{MAX}$, both for all $i,j$ with $j\ne i$, we can make $q(t_{1x},\ldots,t_{Gx}|s_i)=\prod_j^G q(t_{jx}|s_i)$. Then, the \textit{total correlation} term reaches its minimum value $0$ (proved in the extended paper).

Apparently, we have the loss function for multi-\genre\ disentanglement:
\begin{equation}\label{eq:mgd}
\small
\mbox{$L_{m} = \sum_i\sum_{j,j\ne i}\Big[\mathcal H(p(t_i|s_i)) - \mathcal H(p(t_j|s_i))- \mathcal H(p(t_j|\tilde s_i))\Big]$,}
\end{equation}
where $\tilde s_i$ is a sample from the distribution of style \type\ $t_i$.

\begin{table*}[!t]
\centering
\begin{small}
\begin{tabular}{|c|c|c|c|c|c|c|}
\hline
  \multicolumn{2}{|c|}{}& \multicolumn{2}{c|}{Yelp} & \multicolumn{3}{c|}{Amazon}   \\
\cline{3-7}
  \multicolumn{2}{|c|}{}&\newcite{john-etal-2019-disentangled} & Our method&\newcite{john-etal-2019-disentangled} &\multicolumn{2}{c|}{Our method}\\
\hline
   \multicolumn{2}{|c|}{Style \Genre} &\multicolumn{2}{c|}{Sentiment}&\multicolumn{2}{c|}{Sentiment}& Tense\\
   \hline
   \multicolumn{2}{|c|}{Random Guess} &\multicolumn{2}{c|}{61.30}&\multicolumn{2}{c|}{50.82}& 52.30\\
\hline
 \multirow{3}{*}{Vanilla} & Style$^\uparrow$ &\textbf{97.40}&97.31&82.10&\textbf{83.15}&92.50\\
\cline{2-7}
 &Content$^\downarrow$  &65.80&\textbf{65.48}&67.50&\textbf{52.30} &64.40\\
\cline{2-7}
 & Style + Content$^\uparrow$ &\textbf{97.40}&97.31&81.90&\textbf{83.15}&92.50\\
\hline
 \multirow{3}{*}{VAE} &Style$^\uparrow$  &\textbf{97.40}&97.37&81.00&\textbf{81.75}&92.40\\
\cline{2-7}
  &Content$^\downarrow$  &69.70&\textbf{63.02}&69.30&\textbf{52.30}&64.40\\
\cline{2-7}
  &  Style + Content$^\uparrow$&\textbf{97.40}&97.38&81.00&\textbf{81.75}&92.40\\
\hline
\end{tabular}
\end{small}
\smallskip 
\caption{The classification accuracies on each space. The up arrow means a good result should have a larger value, while the down arrow means lower is better. Note that the accuracy for content space is the lower the better, because the goal of disentanglement is to let the content space  not contain any information of style.
For the tense style, there are no such evaluations in previous work, so we did not list any baseline results in this table.}
\label{tab:de}
\end{table*}%

\subsection{Training and Inference}
In the training phase, we minimize the following objective:
\begin{equation}\label{eq:total}
    J=J_\text{AE}+\lambda_aL_\text{al}+\lambda_cL_\text{cl}+\lambda_sL_\text{sc} +\lambda_mL_{m}.
\end{equation}
The item $J_\text{AE}$ can be replaced by $J_\text{VAE}$ for variational autoencoder architectures. $\lambda_a, \lambda_c, \lambda_s$, and $\lambda_m$ in Eq.~\ref{eq:total} are predefined hyperparameters. 

In the inference phase, if we would like to change the current style \type\ to another style \type\ (within \genre\ $i$), we need to follow subsequent steps. First, encode the sentence to a latent representation $h$, and split $h$ to obtain $s_1,\ldots,s_G$ and $c$. Second, we sample a new style vector $\tilde s_i$ from the target style distribution. Finally, replace $s_i$ with $\tilde s_i$ and generate a sentence in the target style by the decoder. 

\section{Experiments}\label{sec:exp}

In this section, we answer the following three  questions: (1)~Is the disentanglement effect comparable to previous works? (2) Is the style-transfer performance comparable to or does it even outperform previous works? (3) Will the proposed method alleviate the training bias problem?

\subsection{Data and Preprocessing}
We tested our method on two datasets: Yelp Service Reviews\footnote{\url{https://github.com/shentianxiao/language-style-transfer}}~\cite{Shen2017Style,Zhao2018Adversarially} and  Amazon Product Reviews\footnote{\url{https://github.com/fuzhenxin/textstyletransferdata}}~\cite{Fu2018Style}. 
These datasets all contain different sentiment \types\ (positive and negative).  
We annotated the  tense label~\footnote{The tense label files are available at: \url{https://drive.google.com/drive/folders/1I1hOZChFPFc2LYreWp1W6l4WpdS0pnVK?usp=sharing}} in the Amazon dataset by matching the sentence tokens with a time word corpus, which is collected from the  TimeBank   dataset\footnote{\url{http://timeml.org}}~\cite{pustejovsky2003timebank}. This is consistent with the method described in \newcite{hu2017toward}. We do not label the Yelp dataset in the same way, because the time information is so vague that our automatic annotation will cause too many errors and  mislead our research.


 \begin{figure*}[!t]
\centering
\resizebox{\linewidth}{!}{
\begin{tabular}{cccc}
 \multicolumn{2}{c}{-----------Yelp-----------} & \multicolumn{2}{c}{-----------Amazon-----------}\\
\includegraphics[width=0.25\textwidth]{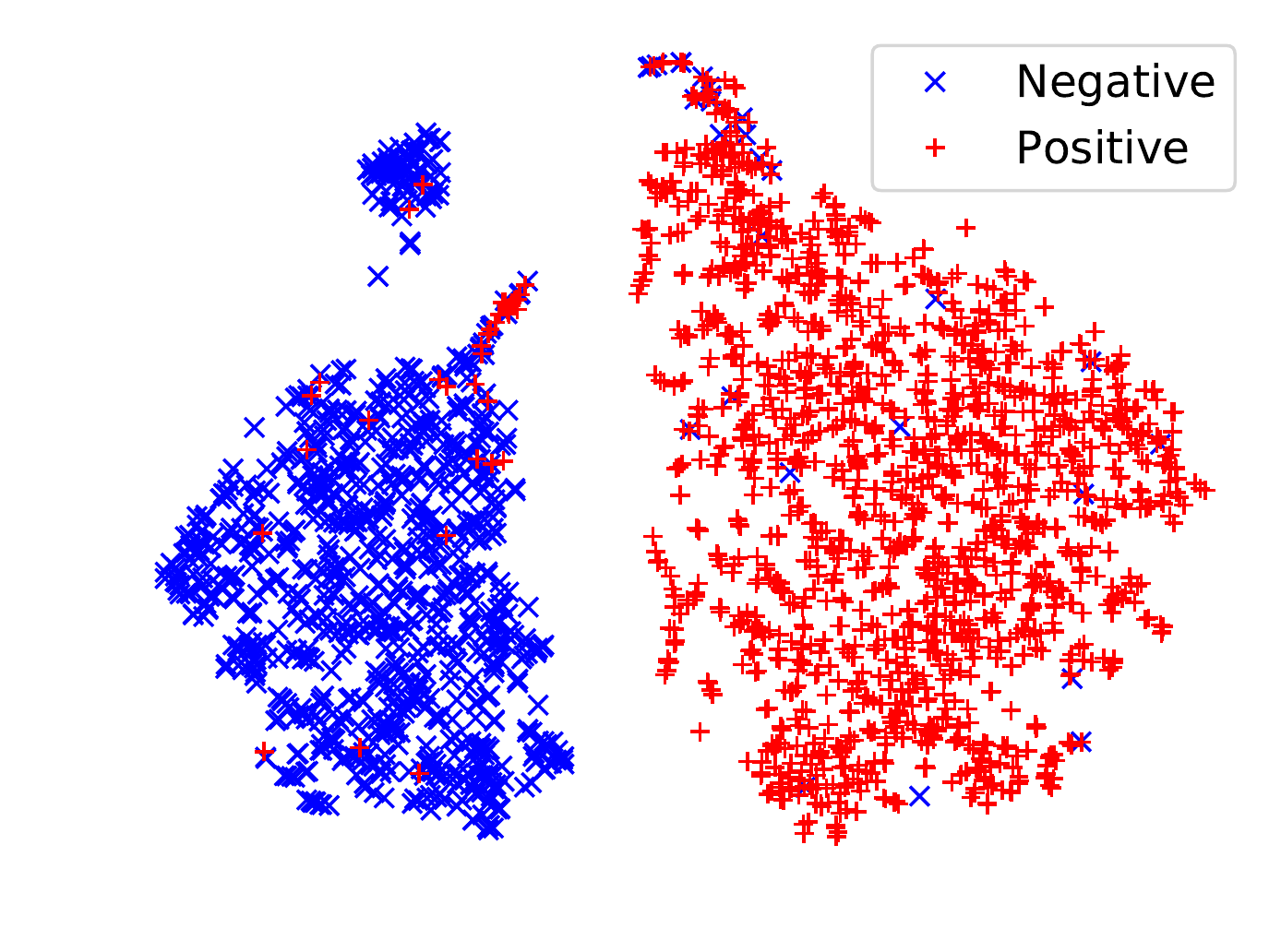}&\includegraphics[width=0.25\textwidth]{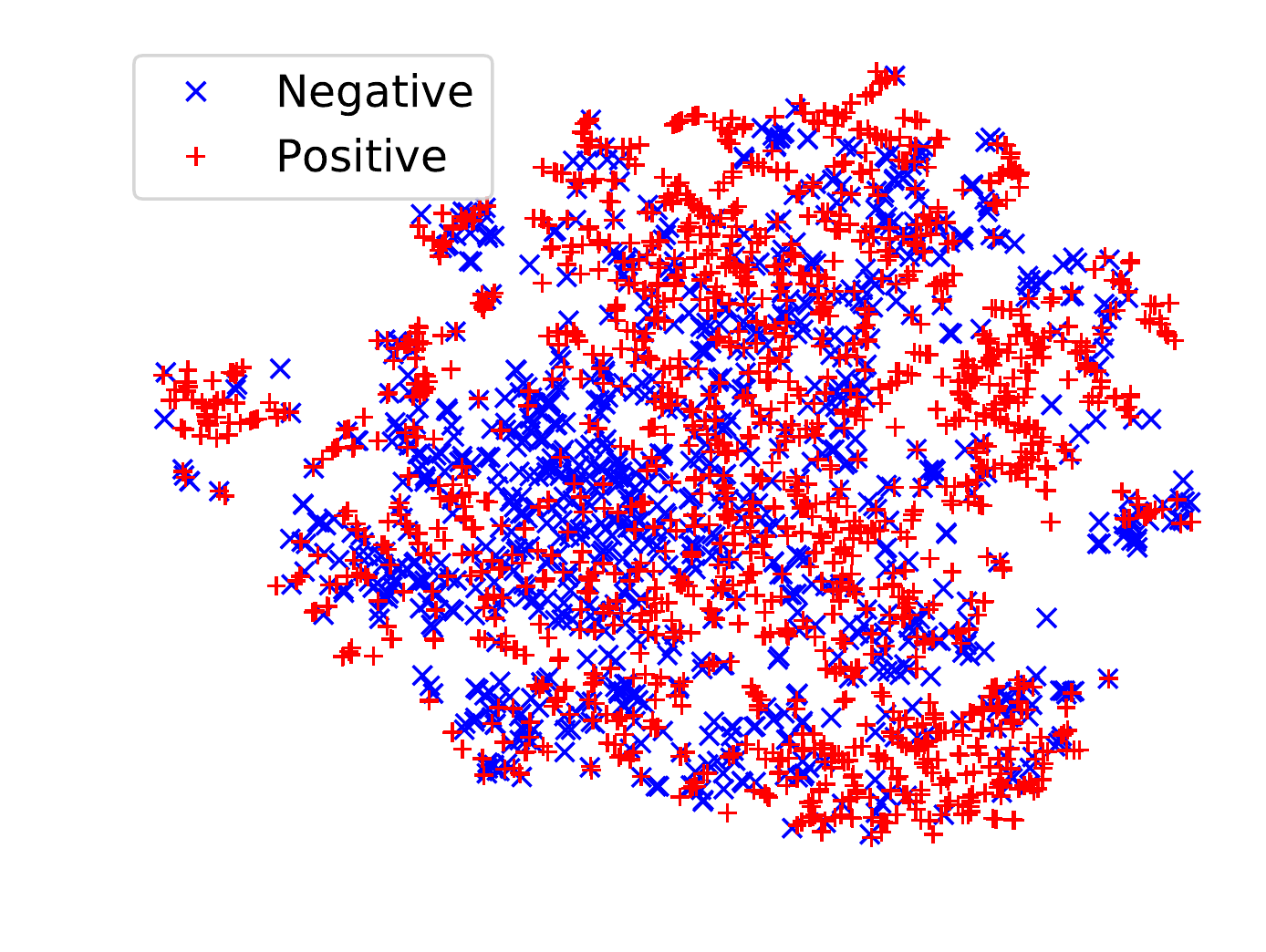}&\includegraphics[width=0.25\textwidth]{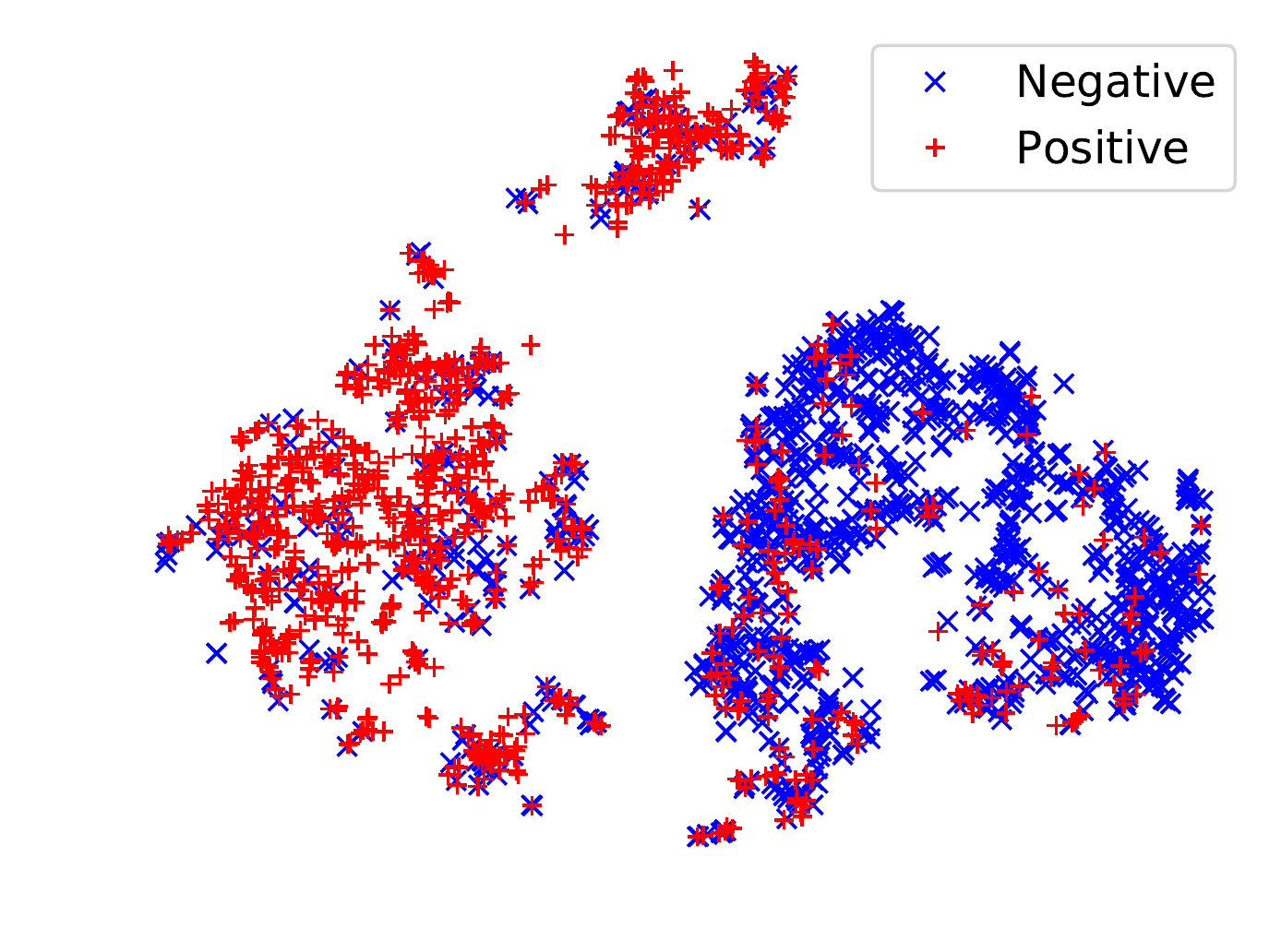}&\includegraphics[width=0.25\textwidth]{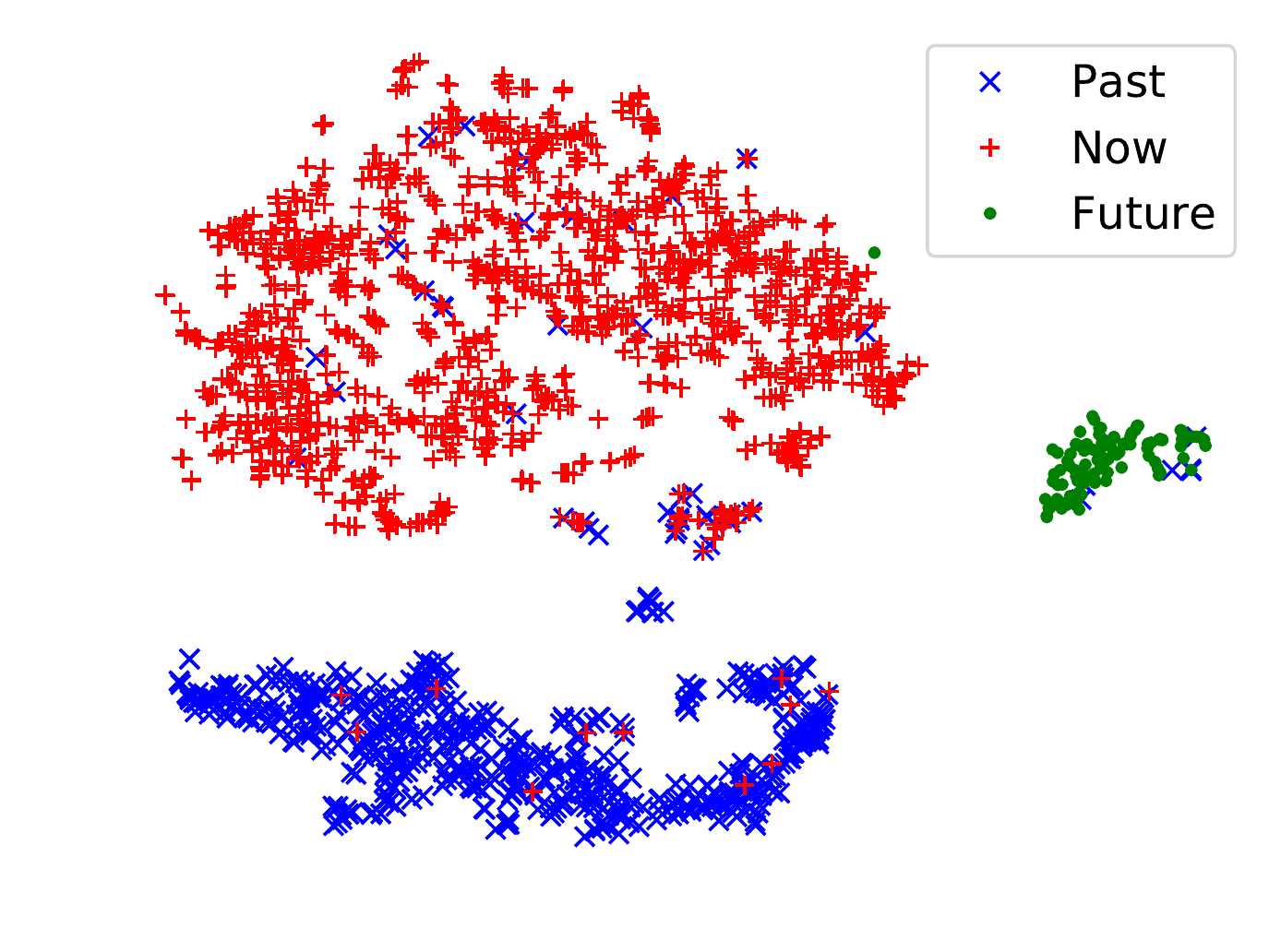}\\
 (a) Vanilla, Sentiment&(b) Vanilla, Content &(c) Vanilla, Sentiment &(d) Vanilla, Tense\\
\includegraphics[width=0.25\textwidth]{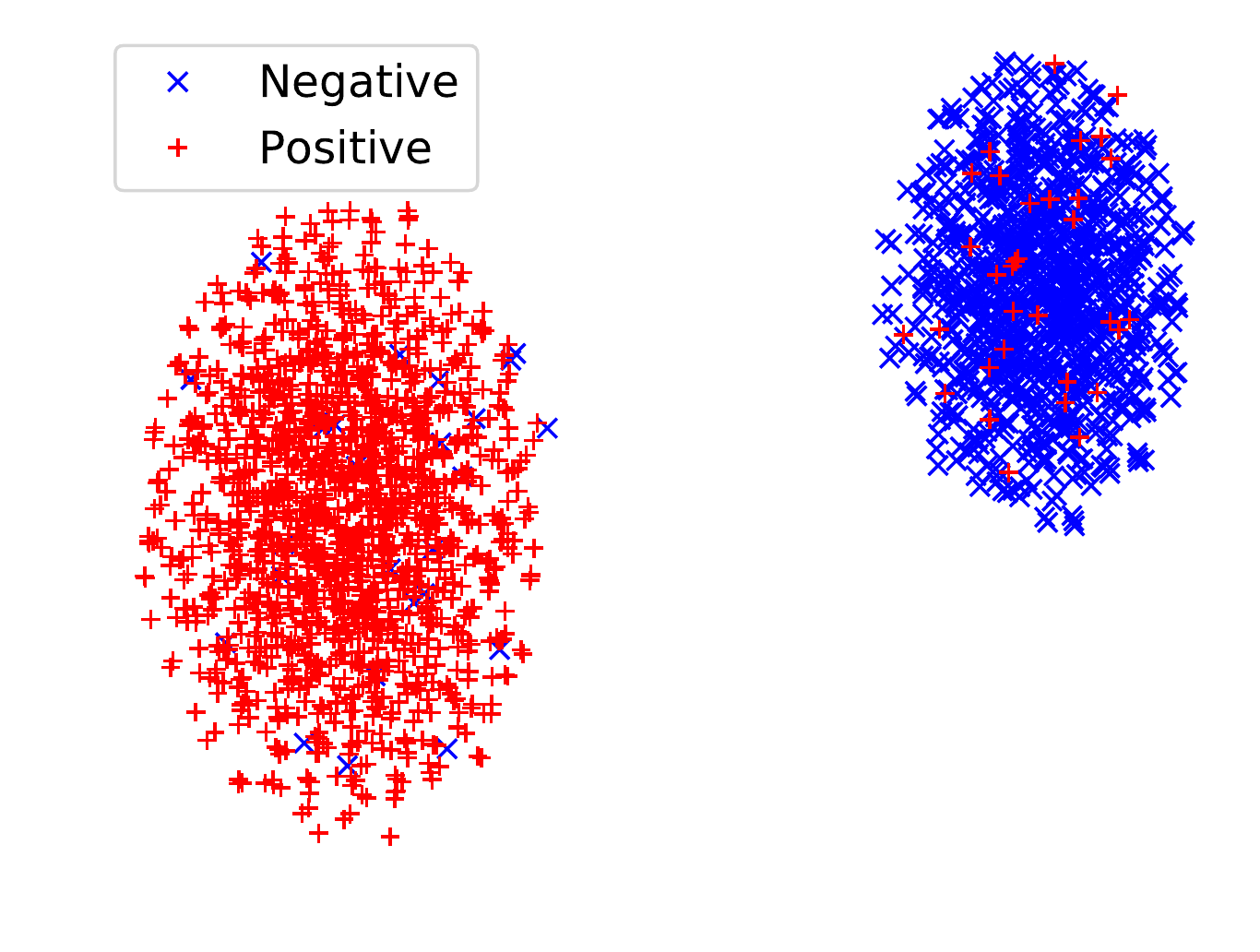}&\includegraphics[width=0.25\textwidth]{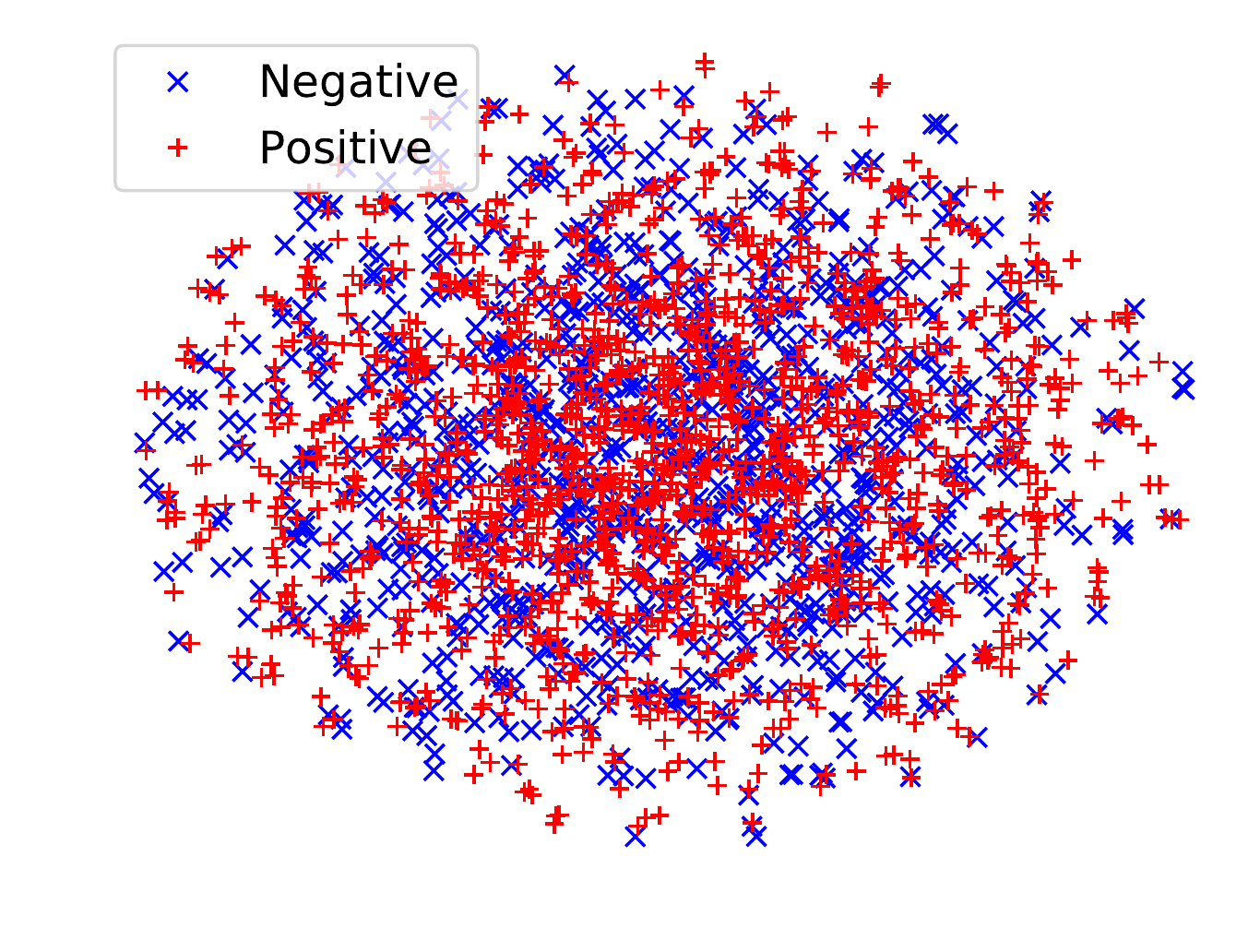}&\includegraphics[width=0.25\textwidth]{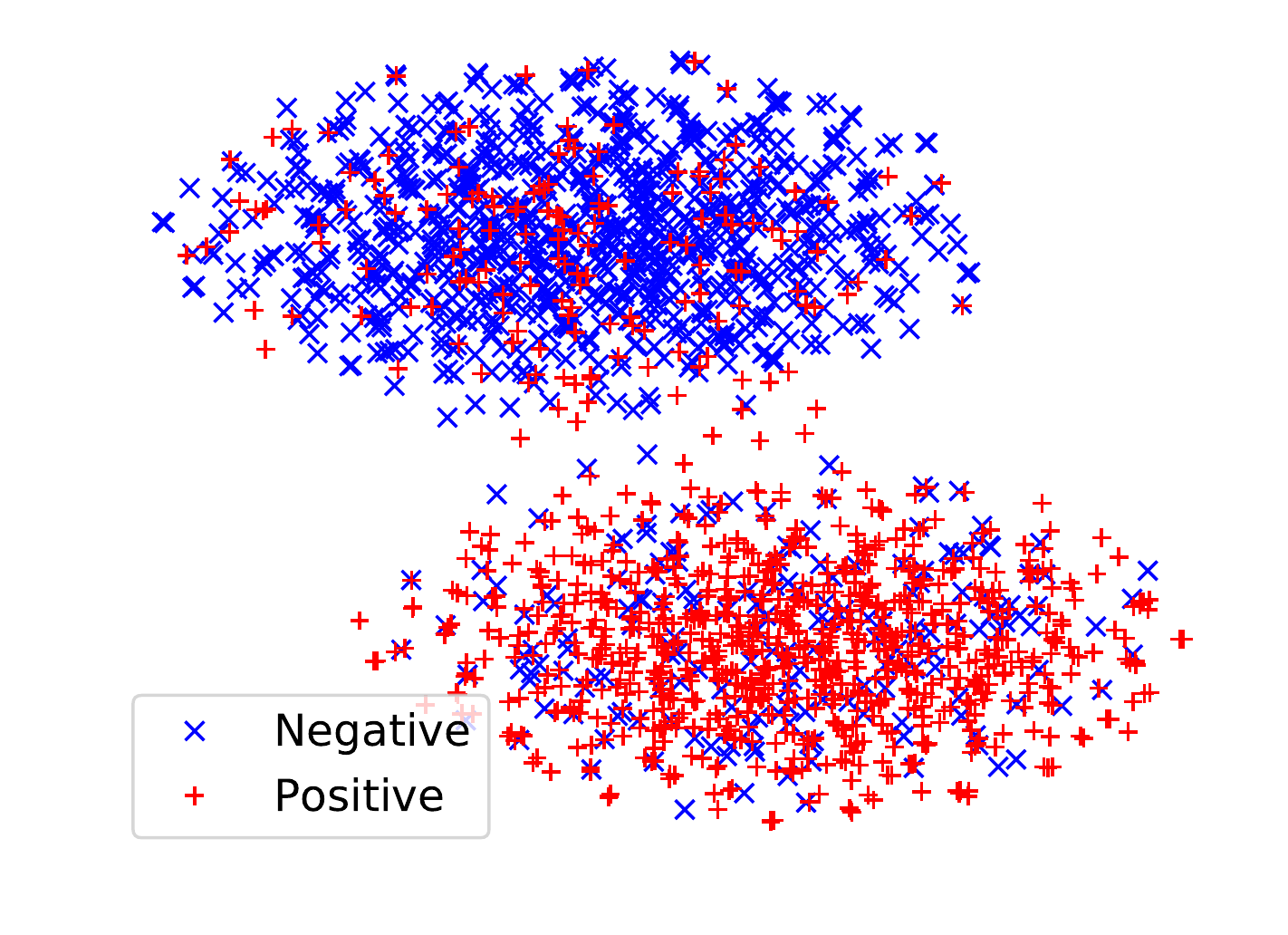}&\includegraphics[width=0.25\textwidth]{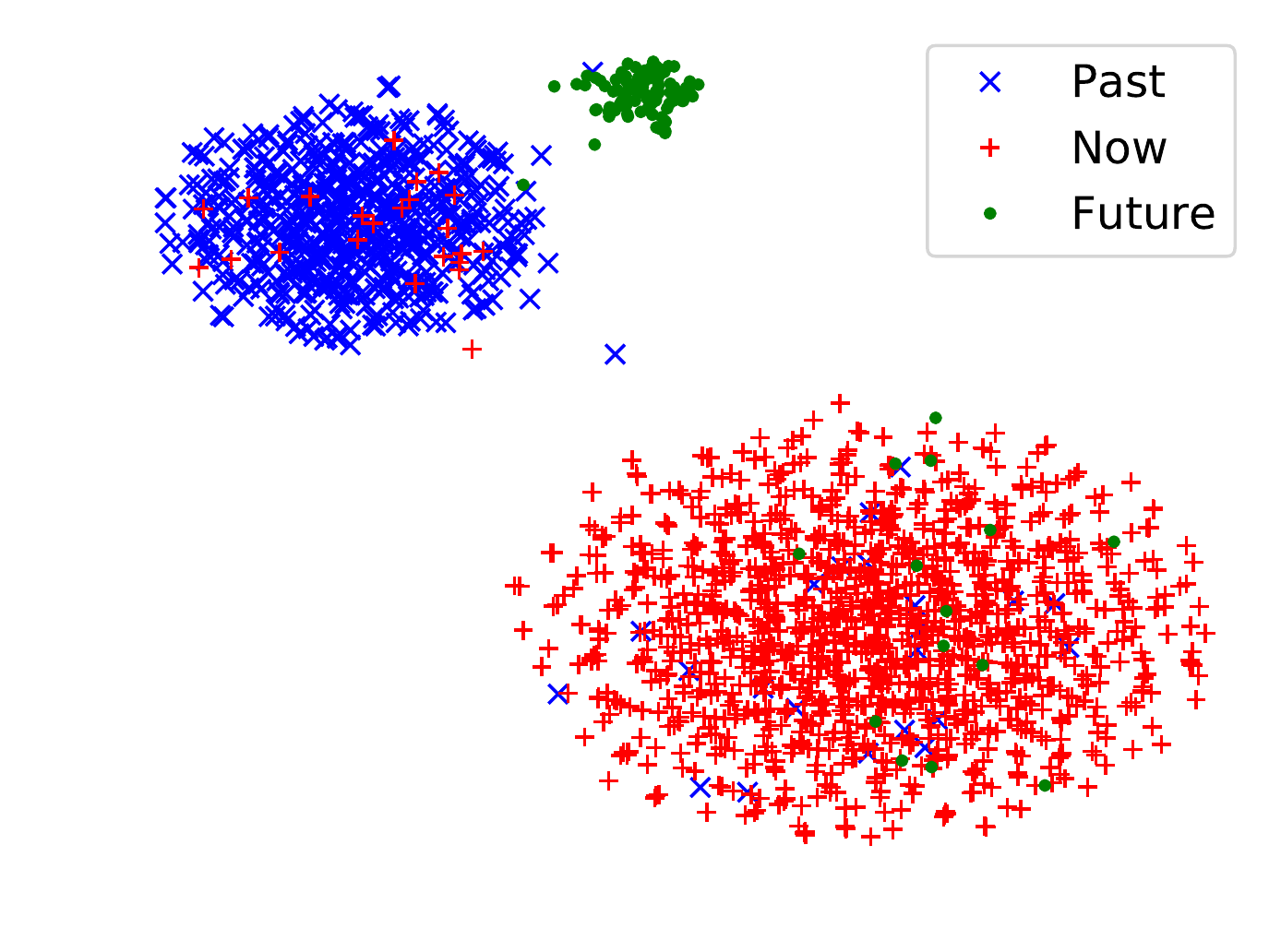}\\
 (e) VAE, Sentiment&(f) VAE, Content &(g) VAE, Sentiment&(h) VAE, Tense\\
\end{tabular}
}

\smallskip 
\caption{The t-SNE visualization of each latent space. (a), (b), (e), and (f) are the style space (``$s$'') and content space (``$c$'') of the Yelp dataset.  (c), (d), (g), and (h) are the two style spaces (sentiment ``$s_1$'' and tense ``$s_2$'') of the Amazon dataset, which is generated simultaneously  in multi-\genre\ disentanglement. All latent spaces are generated by two different methods: a vanilla autoencoder and a variational autoencoder.}
\label{fig:vis}
\end{figure*}

\subsection{Disentanglement Effect}
We use two metrics to evaluate the disentanglement effect. In the first metric, we train separate logistic regression classifiers for all the generated style vectors and content vectors. We compare the performance with previous work in Table~\ref{tab:de}.

According to Table~\ref{tab:de}, we found that the high accuracies of style vectors on Yelp and Amazon sentiment/tense are nearly identical to the corresponding accuracy of competent space (style and content), which means that the style vectors contain nearly all sentiment/tense information. Comparatively, the performance of content vectors are only comparable to a random guess result. These results indicate that our disentanglement method is very effective.


In Fig.~\ref{fig:vis}, we listed the visualization result of each latent space using t-SNE\footnote{In practice, we use the multi-core version for acceleration: \url{https://github.com/DmitryUlyanov/Multicore-TSNE}}~\cite{maaten2008visualizing}. 
For the vanilla auto-encoders, we can see that in (a), (c), and (d), the different style \types, such as positive/negative of sentiment and past/now/future of tense,  apparently stay in separate places of themselves. The margins between any two style \types\ are  sufficient for discrimination. In Fig.~\ref{fig:vis} (b), the content space is indistinguishable due to its lack of sentiment information. 
For the variational autoencoders, in Fig.~\ref{fig:vis} (e), (f), (g), and (h),   we found that all of them are in very regular shape. All disentangled groups of style \types, including positive/negative of sentiment and past/now/future of tense, appear as separate ellipses. The reason for this phenomenon is that we force the KL divergence of each style \type\ to be smaller, but at the same time, we force to discriminate different style \types. Therefore, different style \types\ tend to stay closer but with a large margin between them.

\subsection{Style-Transfer Performance}

To be consistent with previous works, we use four evaluation metrics to evaluate the style-transfer effect of our disentanglement method, and list all the results in Table~\ref{tab:overall}. 

(1) Style-transfer accuracy (STA): we trained two external sentence classifiers for sentiment and tense using TextCNN~\cite{kim2014convolutional} following previous works~\cite{hu2017toward,Fu2018Style,john-etal-2019-disentangled},  and then used them to measure the sentiment/tense accuracy for the style-transferred sentences with the target style \type\ as ground-truth. The external classifier achieves an acceptable accuracy on the validation set (97.68\% on Yelp, 82.3\% on Amazon Sentiment, and 96.5\% on Amazon Tense), which provides a reliable approximation for the sentiment/tense accuracy. 

\begin{table*}[htp]
\centerline{
	\resizebox{\linewidth}{!}{
\begin{tabular}{|l|c|c|c|c| c|c|c |c|c|c|c|}
\hline
    & \multicolumn{5}{c|}{Yelp}                            &\multicolumn{6}{c|}{Amazon}    \\\cline{2-12}
          & STA & CS & WO     &PPL    & BLEU    & STA/Sentiment &STA/Tense &  CS & WO &   PPL & BLEU   \\\hline
 \newcite{Zhao2018Adversarially}      & 0.818& 0.883 &0.272 &     85   &    -     & 0.552&  -  &0.926 &0.169 & 75 & - \\\hline
 \newcite{li-etal-2018-delete}    & 0.862 &\textbf{0.941} &0.522 &    70     &  -     & \color{gray}{0.430} &   -&\color{gray}{\textbf{0.976}} &\color{gray}{\textbf{0.799}}& \color{gray}{65}&- \\\hline
 \newcite{xu2018unpaired}              & 0.803 &0.924 &0.427 &     470    &   -     & 0.723 &  -&0.912 &0.222 & 332&- \\\hline
 Logeswaran et al.~\shortcite{logeswaran2018content}&  0.905&0.879     &0.503     &133  &  17.4      & 0.857  &  0.942 &0.908   &0.356 & 187& 16.6\\\hline
\newcite{lample2019multipleattribute}&  0.877    & 0.856    &0.459     & 48&   14.6& 0.896   & 0.965 &0.897   &0.372 & 92& 18.7\\\hline
 \newcite{john-etal-2019-disentangled}~(Vanilla)& 0.883 &0.915 &0.549 & 52 & 18.7   & 0.720&   0.926 &0.921 &0.354& 73& 16.5\\\hline
 \newcite{john-etal-2019-disentangled}~(VAE)    & 0.934 &0.904 &0.473&   32  & 17.9      & 0.822 &  0.945& 0.900 &0.196&  63&9.8 \\\hline
 \hline
 Ours~(Vanilla)      & 0.877 & 0.908 & \textbf{0.554 } &    45       &   16.1        & 0.789 &    0.963& \textbf{0.930} & \textbf{0.387}& 68& 15.4 \\\hline
 Ours~(Vanilla)$-L_\text{cl}$       & 0.605 & 0.852 & 0.419 &  51    &   13.6    & 0.679  &   0.908& 0.896 & 0.382 &  74& 12.5\\\hline
 Ours~(Vanilla)$-L_\text{sc}$     & \color{gray}{0.383} & \color{gray}{0.875} & \color{gray}{0.550} &   \color{gray}{42}   & 15.8   & 0.746 &    0.881& 0.904 & \textbf{0.387}&  65& 16.9\\\hline
 Ours~(VAE)       & \textbf{0.944} & 0.912 & 0.455  &    27  &  \textbf{21.2}  & \textbf{0.902} &  \textbf{0.993}& 0.900 & 0.338&    44&\textbf{20.1}  \\\hline
 Ours~(VAE)$-L_\text{cl}$      & 0.734 & 0.860 & 0.391 &   32  &  15.8   & 0.751 &   0.932& 0.907 & 0.310 &   57& 13.4 \\\hline
 Ours~(VAE)$-L_\text{sc}$      & 0.656 & 0.868 & 0.438 &  \textbf{25}  & 19.8    & 0.720 &   0.877& 0.898 & 0.345& \textbf{43}&17.8 \\\hline
\end{tabular}
}}

\smallskip 
\caption{The overall style transfer performance. For tense style, we do not have previous works to compare with, so we only listed our own result. STA: style transfer accuracy, CS: cosine similarity w/o sentiment/tense word, WO: word overlap w/o sentiment/tense word. For the sentiment \genre, the transfer direction  is  ``Neg$\rightarrow$Pos'' and ``Pos$\rightarrow$Neg''. For the tense \genre, the transfer direction  is that ``Past$\rightarrow$Now'', ``Now$\rightarrow$Future'', and ``Future$\rightarrow$Past''. Note that the grey numbers have very low STA, which means it always fails in style transferring. So, we highlight the best values of the evaluations apart from them.  }
\label{tab:overall}
\end{table*}%

(2) Cosine-similarity (CS): We computed the cos-similarity between the original sentence's vector and the style-transferred sentence's vector. Each sentence vector is obtained by concatenating the \textit{max}, \textit{min}, \textit{mean} of word vectors after deleting the sentiment/tense words. This is again consistent with previous works~\cite{Fu2018Style,john-etal-2019-disentangled}. The goal of this metric is to evaluate the semantic similarity between the original sentence and the style-transferred sentence apart from sentiment words. 

(3) Word overlap (WO): Following \newcite{john-etal-2019-disentangled}, we calculated the unigram overlap between the original and the style-transferred sentence,  which is defined as the ratio between the number of words in the intersection set and the number of words in the union set of the two sentences.
 
(4) Perplexity (PPL): We applied a trigram KneserNey~\cite{kneser1995improved} language model as the perplexity evaluator. We trained two language models separately on respective datasets, and use the trained model to evaluate the fluency of generated sentences. A lower PPL value represents a more fluent sentence.  

(5) BLEU: We calculate the BLEU 1$\sim$4 score between the original sentence and the style-transferred sentence, and take the average of   BLEU 1$\sim$4 as the BLEU score. Again, we need to delete the sentiment/tense words before evaluation.

By Table~\ref{tab:overall}, in  Yelp and Amazon sentiment, our method is comparable with previous works on the CS and WO metrics except for \citep{li-etal-2018-delete}. But our variational autoencoder (VAE) architectures can outperform \citep{li-etal-2018-delete} by a large margin on the STA metric. Also, our VAE architectures can outperform all the previous methods in STA on the two datasets (with Student's t-test, $p<0.01$). In the Amazon tense experiment, although we do not have previous works to compare with, we achieved a  high STA value, while the CS and WO remained comparable with the same metric in Amazon sentiment. We found that the CS and WO value of vanilla autoencoders (AEs) are higher than the same metrics of VAEs, because VAEs are more flexible in the latent space,  so that the generated sentence is much more varied than vanilla AEs.

For ablation tests, we remove  $L_\text{cl}$ and $L_\text{sc}$ from the objective for comparison. We did not try to remove $L_\text{al}$, because the whole model would become disconnected by doing that. According to Table~\ref{tab:overall}, after $L_\text{cl}$ is removed, the value of STA decreased a lot, because the model can no longer  discriminate different style \types\ without $L_\text{cl}$. 
On the other hand, when we remove $L_\text{sc}$, the STA value gets even lower. This is likely because the content space is flooded with style information without style-content disentanglement; thus, the sentence cannot be completely transferred to the target style.

\begin{table}[!t]
\centerline{
\resizebox{0.8\linewidth}{!}{
\begin{tabular}[t]{|l|c|c|c|}
\hline
 &  & Sentiment  & Tense \\
\hline
 Origin   &STA  &0.8150 &0.9715\\ 
\hline
\multirow{2}{*}{Vanilla} &STA$_\text{keep}$        &0.7170  &0.9220\\
         &$\Delta$   &\textbf{0.0980}  &\textbf{0.0495}\\
\hline
\multirow{2}{*}{Vanilla - $L_{m}$} &STA$_\text{keep}$        &0.7020 &0.6920\\
          &$\Delta$   &    0.1130 &0.2795\\
\hline
\multirow{2}{*}{VAE} &STA$_\text{keep}$        &0.7850  &0.8825\\
      &$\Delta$  & \textbf{0.0300}    & \textbf{0.0890}  \\
\hline
\multirow{2}{*}{VAE - $L_{m}$} &STA$_\text{keep}$        & 0.7260 & 0.8220\\
         &$\Delta$     &0.0890  &0.1495\\
\hline
\end{tabular}
}}

\smallskip 
\caption{\textbf{STA$_\text{keep}$} stands for the current style \genre's STA after another \genre\ is transferred, i.e., we observe the STA of \textit{sentiment} when we are transferring the sentence's \textit{tense}. Also, we observe the STA of \textit{tense} when we are transferring   \textit{sentiment}. $\Delta=\text{STA}-\text{STA}_\text{keep}$.  The line of ``Origin'' represents the CNN predicted result (STA) of the original sentence. }
\label{tab:delta} 
\end{table}%

\subsection{Alleviation of Training Bias}
When the training bias is alleviated, the \type\ of a  style \genre\ tends not to be affected by the transfer of another style \genre's \type.
So, we are trying to measure the style preservation for a baseline style \genre\ (that is held constant) while trying to transfer the other style \genre. Therefore, a lower deviation of STA from the original style \genre\  is better, and removing the loss $L_{m}$ should produce poorer results.

The effect of our multi-\genre\ loss function on alleviating training bias is shown in Table~\ref{tab:delta}, where we listed the model's accuracies on one style \genre\ when the other style \genre\ is transferred (STA$_\text{keep}$ score). We report the STA$_\text{keep}$ of vanilla AEs and VAEs on the two \genres\ (sentiment and tense), and we remove the loss $L_{m}$ from the models for comparison. We found that the STA$_\text{keep}$ score of our full model can be very close to the original sentence's STA score. But if we remove the $L_{m}$ item, the accuracy of any \genre\ would decrease a lot after the style \type\ of another \genre\ is transferred. This fact illustrates the effectiveness of our multi-\genre\ disentanglement method.

\subsection{Human Evaluation}
We conducted a human evaluation on the Yelp dataset and the Amazon dataset, like previous works. We randomly sampled 1,000 cases from the  sentences generated by each model and asked 6 data graders to give each case a sentiment label or tense label (for transfer accuracy (TA)) and two scores on content preservation (CP), and language quality (LQ). Each score is between 1 to 5. The detailed annotation principles are listed in the extended paper. We randomly shuffled the generated sentences to conduct the grading process  in a strictly blind fashion. The human evaluation results are shown in Table~\ref{tab:he}. Our  measure of inter-rater agreements (the Krippendorff's alpha values~(\citeyear{krippendorff2004content})) are also listed in Table~\ref{tab:he}, all of them are acceptable due to Krippendorff's principle~(\citeyear{krippendorff2004content}).
\begin{table}[!t] 
\centerline{
\resizebox{\linewidth}{!}{
\begin{tabular}{|l|l|c|c|c|c|c|c|}
\hline
  && \multicolumn{3}{c|}{Sentiment}& \multicolumn{3}{c|}{Tense}\\\cline{3-8}
 && TA & CP & LQ& TA & CP & LQ\\\hline
\multirow{6}{*}{Yelp} &\newcite{Zhao2018Adversarially} &75.42 &3.23& 3.86 & - & -& -\\\cline{2-8}
&\newcite{john-etal-2019-disentangled}~(Vanilla) &82.11& 3.52 &4.02& - & -& -\\\cline{2-8}
&\newcite{john-etal-2019-disentangled}~(VAE) &85.70 &3.70 &4.26& - & -& -\\\cline{2-8}
&Ours (Vanilla) &84.28 & 3.69& 4.32& - & -& -\\\cline{2-8}
&Ours (VAE) & \textbf{86.04}& \textbf{3.78}&\textbf{4.39}& - & -& -\\\cline{2-8}
& IRA & 0.75 & 0.71& 0.82& - & -& -\\\cline{2-8}
%
\hline
\hline
\multirow{5}{*}{Amazon} &\newcite{john-etal-2019-disentangled}~(Vanilla) &76.35& 3.01 &3.65&87.34 &2.97 & 3.96\\\cline{2-8}
&\newcite{john-etal-2019-disentangled}~(VAE) &79.60 &3.26 &3.76& 88.92&3.14 & 4.14\\\cline{2-8}
&Ours (Vanilla) &79.03 & 3.34& 3.74&91.09 & 3.21&4.09 \\\cline{2-8}
&Ours (VAE) & \textbf{83.28}& \textbf{3.52}&\textbf{4.08}&\textbf{93.45} &\textbf{3.58} & \textbf{4.23}\\\cline{2-8}
& IRA & 0.82 & 0.78& 0.85& 0.93& 0.89& 0.87\\\hline
\end{tabular}
}}
\smallskip 
\caption{Human evaluation results on the Yelp and Amazon dataset. Here, IRA represents inter-rater agreements.}
\label{tab:he}
\end{table}%

%

\section{Related Work}\label{sec:rel}
Disentanglement is a very important method for interpretable deep learning models~\cite{chen2019looks,sha2020estimating,sha2021learn}.
Disentanglement works can be split into implicit disentanglement and explicit disentanglement. We summarize their characteristics as follows, focusing on the sentiment and the tense style for this comparison with previous works.
 
Implicit disentanglement means to disentangle meaningful factors from a variational space,  but we are not sure how many disentangled components  the latent space will be separated. $\beta$-VAEs~\cite{higgins2017beta} and $\beta$-TCVAE~\cite{chen2018isolating} are unsupervised methods that extend variational autoencoders (VAEs)~\cite{kingma2014auto,rezende2014stochastic} and learn disentangled representations by putting a penalty on the \textit{total correlation (TC)} item.
There are also a number of methods that extend $\beta$-VAEs and analyze different variants under specific problems~\cite{moyer2018invariant,mathieu2018disentangling,kumar2017variational,esmaeili2018structured,hoffman2016elbo,narayanaswamy2017learning,kim2018disentangling,rezende2018taming,shao2020controlvae}. However, in the training process of implicit disentanglement methods, some components may be pruned~\cite{stuhmer2019independent}, which may lead to incorrect interpretations of the data.


In comparison, an explicit disentanglement is able to separate the latent space into more interpretable components and control them using latent variables. For example, \newcite{chen2016infogan} present a GAN-based method that maximizes the mutual information between a scalar variable and a generator. Then, the scalar variable can be taken as a controller for the style of the generated text or image. Basically, adversarial methods are always used to guarantee that different disentangled factors are independent~\cite{john-etal-2019-disentangled,romanov2019adversarial}. However, adversarial methods suffered from  oscillating and unstable model parameters, which makes the training hard to converge. Also, 
some research~\cite{elazar-goldberg-2018-adversarial,moyer2018invariant} pointed out that adversarial training is not so reliable in disentanglement-invariant representations.  

One of the most common applications of disentanglement is  text/image style transfer. Apart from some disentangle-free approaches~\cite{preoctiuc2018user,logeswaran2018content,dai-etal-2019-style,lample2019multipleattribute,dankers2019modelling}, there are three ways for style transfer as follows: (1) example imitating: to extract a high-level feature and force the input text/image's feature to approach the example's feature~\cite{gatys2016image}; (2) scalar variable tuning: after disentangling the input into latent scalar variables, slightly tune a scalar variable larger or smaller, expecting the generated text/image would change accordingly~\cite{chen2016infogan,hu2017toward,kumar2017variational,malandrakis-etal-2019-controlled}; and (3) vector-variable replacing: after disentangling the input sentence, sample some target style's vectors and replace the original style vector with their average~\cite{john-etal-2019-disentangled}. 
However, an averaged style vector usually means that we have to sample some examples in the target style and calculate the average of their style vector, which is inconvenient compared to our unified style representation.

Apart from sentiment and tense, there are also other kinds of styles. \newcite{kang2019xslue} proposed a benchmark for style classification and discussed various kinds of styles, including \textit{emotion}, \textit{age}, and \textit{formality}. \newcite{kang2019male} proposed a parallel persona style dataset.

\section{Conclusion}\label{sec:con}
In this paper, we proposed a unified distribution method as well as multiple loss functions to avoid adversarial training in the disentangling process. 
Our method is easy to be applied in multi-type disentanglement. we conducted style disentanglement experiments and style transfer experiments to prove the effectiveness of our method.

\section*{Acknowledgments}
This work was supported by the EPSRC grant ``Unlocking the Potential of AI for English Law'', a JP Morgan PhD Fellowship, the Alan Turing Institute under the EPSRC grant EP/N510129/1, and the AXA Research Fund. We also acknowledge the use of Oxford's Advanced Research Computing (ARC) facility, of the EPSRC-funded Tier 2 facility
JADE (EP/P020275/1), and of GPU computing support by Scan Computers International Ltd.
\small
\bibliography{cite}

\begin{thebibliography}{44}
\providecommand{\natexlab}[1]{#1}
\providecommand{\url}[1]{\texttt{#1}}
\providecommand{\urlprefix}{URL }
\expandafter\ifx\csname urlstyle\endcsname\relax
  \providecommand{\doi}[1]{doi:\discretionary{}{}{}#1}\else
  \providecommand{\doi}{doi:\discretionary{}{}{}\begingroup
  \urlstyle{rm}\Url}\fi

\bibitem[{Chen et~al.(2019)Chen, Li, Tao, Barnett, Rudin, and
  Su}]{chen2019looks}
Chen, C.; Li, O.; Tao, D.; Barnett, A.; Rudin, C.; and Su, J.~K. 2019.
\newblock This Looks Like That: Deep Learning for Interpretable Image
  Recognition.
\newblock In \emph{Advances in Neural Information Processing Systems},
  8930--8941.

\bibitem[{Chen et~al.(2018)Chen, Li, Grosse, and Duvenaud}]{chen2018isolating}
Chen, T.~Q.; Li, X.; Grosse, R.~B.; and Duvenaud, D.~K. 2018.
\newblock Isolating Sources of Disentanglement in Variational Autoencoders.
\newblock In \emph{Advances in Neural Information Processing Systems},
  2610--2620.

\bibitem[{Chen et~al.(2016)Chen, Duan, Houthooft, Schulman, Sutskever, and
  Abbeel}]{chen2016infogan}
Chen, X.; Duan, Y.; Houthooft, R.; Schulman, J.; Sutskever, I.; and Abbeel, P.
  2016.
\newblock Info{GAN}: Interpretable Representation Learning by Information
  Maximizing generative adversarial nets.
\newblock In \emph{Advances in Neural Information Processing Systems},
  2172--2180.

\bibitem[{Dai et~al.(2019)Dai, Liang, Qiu, and Huang}]{dai-etal-2019-style}
Dai, N.; Liang, J.; Qiu, X.; and Huang, X. 2019.
\newblock Style Transformer: Unpaired Text Style Transfer without Disentangled
  Latent Representation.
\newblock In \emph{Proceedings of the 57th Annual Meeting of the Association
  for Computational Linguistics}, 5997--6007. Association for Computational
  Linguistics.

\bibitem[{Dankers et~al.(2019)Dankers, Rei, Lewis, and
  Shutova}]{dankers2019modelling}
Dankers, V.; Rei, M.; Lewis, M.; and Shutova, E. 2019.
\newblock Modelling the Interplay of Metaphor and Emotion Through Multitask
  Learning.
\newblock In \emph{Proceedings of the 2019 Conference on Empirical Methods in
  Natural Language Processing and the 9th International Joint Conference on
  Natural Language Processing (EMNLP-IJCNLP)}, 2218--2229. Association for
  Computational Linguistics.

\bibitem[{Devroye(1996)}]{devroye1996random}
Devroye, L. 1996.
\newblock Random Variate Generation in One Line of Code.
\newblock In \emph{Proceedings of the Winter Simulation Conference}, 265--272.
  IEEE.

\bibitem[{Elazar and Goldberg(2018)}]{elazar-goldberg-2018-adversarial}
Elazar, Y.; and Goldberg, Y. 2018.
\newblock Adversarial Removal of Demographic Attributes from Text Data.
\newblock In \emph{Proceedings of the 2018 Conference on Empirical Methods in
  Natural Language Processing}, 11--21. Association for Computational
  Linguistics.

\bibitem[{Esmaeili et~al.(2018)Esmaeili, Wu, Jain, Bozkurt, Siddharth, Paige,
  Brooks, Dy, and van~de Meent}]{esmaeili2018structured}
Esmaeili, B.; Wu, H.; Jain, S.; Bozkurt, A.; Siddharth, N.; Paige, B.; Brooks,
  D.~H.; Dy, J.; and van~de Meent, J.-W. 2018.
\newblock Structured Disentangled Representations.
\newblock \emph{arXiv preprint arXiv:1804.02086}.

\bibitem[{Fu et~al.(2018)Fu, Tan, Peng, Zhao, and Rui}]{Fu2018Style}
Fu, Z.; Tan, X.; Peng, N.; Zhao, D.; and Rui, Y. 2018.
\newblock Style Transfer in Text: Exploration and Evaluation.
\newblock In \emph{Proceedings of the 32th AAAI Conference on Artificial
  Intelligence}.

\bibitem[{Gatys, Ecker, and Bethge(2016)}]{gatys2016image}
Gatys, L.~A.; Ecker, A.~S.; and Bethge, M. 2016.
\newblock Image Style Transfer Using Convolutional Neural Networks.
\newblock In \emph{Proceedings of the IEEE Conference on Computer Vision and
  Pattern Recognition}, 2414--2423.

\bibitem[{Higgins et~al.(2017)Higgins, Matthey, Pal, Burgess, Glorot,
  Botvinick, Mohamed, and Lerchner}]{higgins2017beta}
Higgins, I.; Matthey, L.; Pal, A.; Burgess, C.; Glorot, X.; Botvinick, M.;
  Mohamed, S.; and Lerchner, A. 2017.
\newblock Beta-VAE: Learning Basic Visual Concepts with a Constrained
  Variational Framework.
\newblock \emph{International Conference on Learning Representations}, 2(5): 6.

\bibitem[{Hochreiter and Schmidhuber(1997)}]{hochreiter1997long}
Hochreiter, S.; and Schmidhuber, J. 1997.
\newblock Long Short-Term Memory.
\newblock \emph{Neural Computation}, 9(8): 1735--1780.

\bibitem[{Hoffman and Johnson(2016)}]{hoffman2016elbo}
Hoffman, M.~D.; and Johnson, M.~J. 2016.
\newblock {ELBO} Surgery: Yet Another Way to Carve Up the Variational Evidence
  Lower Bound.
\newblock In \emph{Proceedings of the Workshop in Advances in Approximate
  Bayesian Inference, NIPS}, volume~1.

\bibitem[{Hu et~al.(2017)Hu, Yang, Liang, Salakhutdinov, and
  Xing}]{hu2017toward}
Hu, Z.; Yang, Z.; Liang, X.; Salakhutdinov, R.; and Xing, E.~P. 2017.
\newblock Toward Controlled Generation of Text.
\newblock In \emph{Proceedings of the 34th International Conference on Machine
  Learning-Volume 70}, 1587--1596. JMLR. org.

\bibitem[{John et~al.(2019)John, Mou, Bahuleyan, and
  Vechtomova}]{john-etal-2019-disentangled}
John, V.; Mou, L.; Bahuleyan, H.; and Vechtomova, O. 2019.
\newblock Disentangled Representation Learning for Non-Parallel Text Style
  Transfer.
\newblock In \emph{Proceedings of the 57th Annual Meeting of the Association
  for Computational Linguistics}, 424--434. Association for Computational
  Linguistics.

\bibitem[{Kang, Gangal, and Hovy(2019)}]{kang2019male}
Kang, D.; Gangal, V.; and Hovy, E. 2019.
\newblock (Male, Bachelor) and (Female, {P}h.{D}) have Different Connotations:
  Parallelly Annotated Stylistic Language Dataset with Multiple Personas.
\newblock In \emph{Proceedings of the 2019 Conference on Empirical Methods in
  Natural Language Processing and the 9th International Joint Conference on
  Natural Language Processing (EMNLP-IJCNLP)}, 1696--1706. Association for
  Computational Linguistics.

\bibitem[{Kang and Hovy(2019)}]{kang2019xslue}
Kang, D.; and Hovy, E. 2019.
\newblock {XSLUE}: A Benchmark and Analysis Platform for Cross-style Language
  Understanding and Evaluation.
\newblock \emph{arXiv preprint arXiv:1911.03663}.

\bibitem[{Kim and Mnih(2018)}]{kim2018disentangling}
Kim, H.; and Mnih, A. 2018.
\newblock Disentangling by Factorising.
\newblock \emph{arXiv preprint arXiv:1802.05983}.

\bibitem[{Kim(2014)}]{kim2014convolutional}
Kim, Y. 2014.
\newblock Convolutional Neural Networks for Sentence Classification.
\newblock \emph{arXiv preprint arXiv:1408.5882}.

\bibitem[{Kingma and Welling(2014)}]{kingma2014auto}
Kingma, D.~P.; and Welling, M. 2014.
\newblock Auto-encoding Variational Bayes.
\newblock \emph{Proceedings of the International Conference on Learning
  Representations}.

\bibitem[{Kneser and Ney(1995)}]{kneser1995improved}
Kneser, R.; and Ney, H. 1995.
\newblock Improved Backing-off for M-gram Language Modeling.
\newblock In \emph{1995 International Conference on Acoustics, Speech, and
  Signal Processing}, volume~1, 181--184. IEEE.

\bibitem[{Krippendorff(2004)}]{krippendorff2004content}
Krippendorff, K. 2004.
\newblock Content Analysis: An Introduction to Its Methodology Thousand Oaks.
\newblock \emph{Calif.: Sage}.

\bibitem[{Kumar, Sattigeri, and Balakrishnan(2017)}]{kumar2017variational}
Kumar, A.; Sattigeri, P.; and Balakrishnan, A. 2017.
\newblock Variational Inference of Disentangled Latent Concepts from Unlabeled
  Observations.
\newblock \emph{arXiv preprint arXiv:1711.00848}.

\bibitem[{Lample et~al.(2019)Lample, Subramanian, Smith, Denoyer, Ranzato, and
  Boureau}]{lample2019multipleattribute}
Lample, G.; Subramanian, S.; Smith, E.; Denoyer, L.; Ranzato, M.; and Boureau,
  Y.-L. 2019.
\newblock Multiple-Attribute Text Rewriting.
\newblock In \emph{Proceedings of the International Conference on Learning
  Representations}.

\bibitem[{Li et~al.(2018)Li, Jia, He, and Liang}]{li-etal-2018-delete}
Li, J.; Jia, R.; He, H.; and Liang, P. 2018.
\newblock Delete, Retrieve, Generate: a Simple Approach to Sentiment and Style
  Transfer.
\newblock In \emph{Proceedings of the 2018 Conference of the North {A}merican
  Chapter of the Association for Computational Linguistics: Human Language
  Technologies, Volume 1 (Long Papers)}, 1865--1874. Association for
  Computational Linguistics.

\bibitem[{Logeswaran, Lee, and Bengio(2018)}]{logeswaran2018content}
Logeswaran, L.; Lee, H.; and Bengio, S. 2018.
\newblock Content Preserving Text Generation with Attribute Controls.
\newblock In \emph{Advances in Neural Information Processing Systems},
  5103--5113.

\bibitem[{Louizos et~al.(2015)Louizos, Swersky, Li, Welling, and
  Zemel}]{louizos2015variational}
Louizos, C.; Swersky, K.; Li, Y.; Welling, M.; and Zemel, R. 2015.
\newblock The Variational Fair Autoencoder.
\newblock \emph{arXiv preprint arXiv:1511.00830}.

\bibitem[{Maaten and Hinton(2008)}]{maaten2008visualizing}
Maaten, L. v.~d.; and Hinton, G. 2008.
\newblock Visualizing Data Using t-SNE.
\newblock \emph{Journal of Machine Learning Research}, 9(Nov): 2579--2605.

\bibitem[{Malandrakis et~al.(2019)Malandrakis, Shen, Goyal, Gao, Sethi, and
  Metallinou}]{malandrakis-etal-2019-controlled}
Malandrakis, N.; Shen, M.; Goyal, A.; Gao, S.; Sethi, A.; and Metallinou, A.
  2019.
\newblock Controlled Text Generation for Data Augmentation in Intelligent
  Artificial Agents.
\newblock In \emph{Proceedings of the 3rd Workshop on Neural Generation and
  Translation}, 90--98. Association for Computational Linguistics.

\bibitem[{Mathieu et~al.(2018)Mathieu, Rainforth, Narayanaswamy, and
  Teh}]{mathieu2018disentangling}
Mathieu, E.; Rainforth, T.; Narayanaswamy, S.; and Teh, Y.~W. 2018.
\newblock Disentangling Disentanglement in Variational Autoencoders.
\newblock \emph{arXiv preprint arXiv:1812.02833}.

\bibitem[{Moyer et~al.(2018)Moyer, Gao, Brekelmans, Galstyan, and
  Ver~Steeg}]{moyer2018invariant}
Moyer, D.; Gao, S.; Brekelmans, R.; Galstyan, A.; and Ver~Steeg, G. 2018.
\newblock Invariant Representations without Adversarial Training.
\newblock In \emph{Advances in Neural Information Processing Systems},
  9084--9093.

\bibitem[{Narayanaswamy et~al.(2017)Narayanaswamy, Paige, Van~de Meent,
  Desmaison, Goodman, Kohli, Wood, and Torr}]{narayanaswamy2017learning}
Narayanaswamy, S.; Paige, T.~B.; Van~de Meent, J.-W.; Desmaison, A.; Goodman,
  N.; Kohli, P.; Wood, F.; and Torr, P. 2017.
\newblock Learning Disentangled Representations with Semi-supervised Deep
  Generative Models.
\newblock In \emph{Advances in Neural Information Processing Systems},
  5925--5935.

\bibitem[{Preo{\c{t}}iuc-Pietro and Ungar(2018)}]{preoctiuc2018user}
Preo{\c{t}}iuc-Pietro, D.; and Ungar, L. 2018.
\newblock User-Level Race and Ethnicity Predictors from {T}witter Text.
\newblock In \emph{Proceedings of the 27th International Conference on
  Computational Linguistics}, 1534--1545. Association for Computational
  Linguistics.

\bibitem[{Pustejovsky et~al.(2003)Pustejovsky, Hanks, Sauri, See, Gaizauskas,
  Setzer, Radev, Sundheim, Day, Ferro et~al.}]{pustejovsky2003timebank}
Pustejovsky, J.; Hanks, P.; Sauri, R.; See, A.; Gaizauskas, R.; Setzer, A.;
  Radev, D.; Sundheim, B.; Day, D.; Ferro, L.; et~al. 2003.
\newblock The Timebank Corpus.
\newblock In \emph{Corpus Linguistics}, volume 2003, 40. Lancaster, UK.

\bibitem[{Rezende, Mohamed, and Wierstra(2014)}]{rezende2014stochastic}
Rezende, D.~J.; Mohamed, S.; and Wierstra, D. 2014.
\newblock Stochastic Backpropagation and Variational Inference in Deep Latent
  Gaussian Models.
\newblock \emph{arXiv preprint arXiv:1401.4082}.

\bibitem[{Rezende and Viola(2018)}]{rezende2018taming}
Rezende, D.~J.; and Viola, F. 2018.
\newblock Taming {VAE}s.
\newblock \emph{arXiv preprint arXiv:1810.00597}.

\bibitem[{Romanov et~al.(2019)Romanov, Rumshisky, Rogers, and
  Donahue}]{romanov2019adversarial}
Romanov, A.; Rumshisky, A.; Rogers, A.; and Donahue, D. 2019.
\newblock Adversarial Decomposition of Text Representation.
\newblock In \emph{Proceedings of the 2019 Conference of the North {A}merican
  Chapter of the Association for Computational Linguistics: Human Language
  Technologies, Volume 1 (Long and Short Papers)}, 815--825. Association for
  Computational Linguistics.

\bibitem[{Sha, Camburu, and Lukasiewicz(2021)}]{sha2021learn}
Sha, L.; Camburu, O.-M.; and Lukasiewicz, T. 2021.
\newblock Learning from the Best: Rationalizing Predictions by Adversarial
  Information Calibration.
\newblock In \emph{Proceedings of the 35th AAAI Conference on Artificial
  Intelligence}.

\bibitem[{Sha et~al.(2020)Sha, Shi, Chen, Zhang, and Wang}]{sha2020estimating}
Sha, L.; Shi, C.; Chen, Q.; Zhang, L.; and Wang, H. 2020.
\newblock Estimating Minimum Operation Steps via Memory-based Recurrent
  Calculation Network.
\newblock In \emph{Proceedings of the 2020 international joint conference on
  neural networks (IJCNN). IEEE}.

\bibitem[{Shao et~al.(2020)Shao, Yao, Sun, Zhang, Liu, Liu, Wang, and
  Abdelzaher}]{shao2020controlvae}
Shao, H.; Yao, S.; Sun, D.; Zhang, A.; Liu, S.; Liu, D.; Wang, J.; and
  Abdelzaher, T. 2020.
\newblock Control{VAE}: Controllable variational autoencoder.
\newblock In \emph{Proceedings of the International Conference on Machine
  Learning}, 8655--8664. PMLR.

\bibitem[{Shen et~al.(2017)Shen, Lei, Barzilay, and Jaakkola}]{Shen2017Style}
Shen, T.; Lei, T.; Barzilay, R.; and Jaakkola, T. 2017.
\newblock Style Transfer from Non-Parallel Text by Cross-Alignment.
\newblock In \emph{Proceedings of the Advances in Neural Information Processing
  Systems}, 6833--6844.

\bibitem[{St{\"u}hmer, Turner, and Nowozin(2019)}]{stuhmer2019independent}
St{\"u}hmer, J.; Turner, R.~E.; and Nowozin, S. 2019.
\newblock Independent Subspace Analysis for Unsupervised Learning of
  Disentangled Representations.
\newblock \emph{arXiv preprint arXiv:1909.05063}.

\bibitem[{Xu et~al.(2018)Xu, Sun, Zeng, Zhang, Ren, Wang, and
  Li}]{xu2018unpaired}
Xu, J.; Sun, X.; Zeng, Q.; Zhang, X.; Ren, X.; Wang, H.; and Li, W. 2018.
\newblock Unpaired Sentiment-to-Sentiment Translation: A Cycled Reinforcement
  Learning Approach.
\newblock In \emph{Proceedings of the 56th Annual Meeting of the Association
  for Computational Linguistics (Volume 1: Long Papers)}, 979--988. Association
  for Computational Linguistics.

\bibitem[{Zhao et~al.(2018)Zhao, Kim, Zhang, Rush, and
  LeCun}]{Zhao2018Adversarially}
Zhao, J.; Kim, Y.; Zhang, K.; Rush, A.~M.; and LeCun, Y. 2018.
\newblock Adversarially Regularized Autoencoders.
\newblock In \emph{Proceedings of the 35th International Conference on Machine
  Learning}, 5897–--5906. JMLR. org.

\end{thebibliography}
 \clearpage
\appendix

\section*{Appendices}
\section{Proofs of the Equations}
\subsection{Proof of Equation~\ref{eq:ict}}\label{ict}
\begin{proof}
\begin{equation}
\begin{small}
\begin{aligned}
I(c,t)&=\mathbb{E}_x\Big[\int_c\sum_{t}p(c,t|x)\log\frac{p(c,t|x)}{p(c|x)p(t)}\Big]\\
&\le\mathbb{E}_x\Big[\int_c\sum_{t}p(c|t,x)(\log p(c|t,x)-\log p(c|x))\Big]\\
&=\mathbb{E}_x\Big[\int_c\sum_{t}(p(c|t,x)\log p(c|t,x)\\
  &\qquad\qquad\qquad-p(c|t,x)\log \sum_{t'}p(c|t',x)p(t'))\Big]\\
&\le\mathbb{E}_x\Big[\int_c\sum_{t}(p(c|t,x)\log p(c|t,x)\\
  &\qquad\qquad\qquad-\sum_{t'}p(t')p(c|t,x)\log p(c|t',x))\Big]\\
&=\mathbb{E}_x\Big[\sum_{t'}p(t')KL(p(c|t,x)||p(c|t',x))\Big]\\
\end{aligned}
\end{small}
\end{equation}
\end{proof}

\subsection{Proof of Equation~\ref{eq:qsss}}\label{qsss}
\begin{proof}
\begin{equation}
\begin{aligned}
q&(s_1,s_2\ldots,s_G)\\
&=\mathbb E_x\Big[\prod_i^Gq(s_i|T_{1x},\ldots,T_{Gx})\Big]\\
&=\sum_x\Big[\prod_i^Gq(s_i|T_{1x},\ldots,T_{Gx})p(x)\Big]\\
&=\sum_x\Big[\frac{\prod_i^Gq(T_{1x},\ldots,T_{Gx}|s_i)q(s_i)}{p(x)^{G-1}}\Big]\\
\end{aligned}
\end{equation}
\end{proof}

\subsection{Proof of $L_{tc}=0$ when irrelevant condition stands}\label{tc0}
\begin{proof}
When the irrelevant condition stands, we have:
\begin{equation}
q(T_{jx}|s_i,T_{kx(k=1\ldots G,k\ne j)})=q(T_{jx}|s_i)
\end{equation}

Then, Equation~\ref{eq:qsss} becomes:
\begin{equation}
\begin{aligned} 
q&(s_1,s_2\ldots,s_G)=\sum_x\Big[\frac{\prod_i^G\prod_j^G q(T_{jx}|s_i)q(s_i)}{p(x)^{G-1}}\Big]\\
&=\sum_x\Big[\frac{\prod_i^G\prod_j^G \big(q(T_{jx}|s_i)q(s_i)\big)}{\Big(\prod_j^Gp(T_{jx})\Big)^{G-1}q(s_i)^{G-1}}\Big]\\
&=\sum_x\Big[\frac{\prod_i^G\prod_j^G \big(q(s_i|T_{jx})\big)\prod_j^Gp(T_{jx})}{q(s_i)^{G-1}}\Big]\\
&=\sum_x\Big[\frac{\prod_i^G q(s_i|T_{ix})q(s_i)^{G-1}\prod_j^Gp(T_{jx})}{q(s_i)^{G-1}}\Big]\\
&=\sum_x\Big[\prod_i^G q(s_i|T_{ix})\prod_j^Gp(T_{jx})\Big]\\
&=\sum_x\Big[\prod_i^G q(s_i,T_{ix})\Big]\\
&=\prod_i^Gq(s_i)\sum_x\Big[\prod_i^G q(T_{ix}|s_i)\Big]\\
&=\prod_i^Gq(s_i)\sum_x\Big[q(T_{1x},\ldots, T_{Gx}|s_i)\Big]\\
&=\prod_i^Gq(s_i)\\
\end{aligned}
\end{equation}
Then,
\begin{equation}
\begin{aligned}  
&L_{tc}=KL(q(s_1,\ldots,s_G)||\prod_iq(s_i))\\
&=\int q(s_1,\ldots,s_G)\log\frac{ q(s_1,\ldots,s_G)}{\prod_iq(s_i)}=0\\
\end{aligned}
\end{equation}

\end{proof}

\section{Proofs of the Theorems}
\subsection{Proof of $I(s,c)$}\label{proof:ics}
\begin{proof}
$I(s,c)$ is defined as follows:
\begin{equation}
I(s,c)=\int\int p(s,c)\log \frac{p(s,c)}{p(s)p(c)}
\end{equation}
Then, we have:
\begin{equation}
\begin{aligned}
I(s,c)&=\int_s\int_c p(s,c)\big(\log p(s,c)-\log (p(s)p(c))\big)\\
 & = \int_s\int_c p(s,c)\big(\log \mathbb E_x[p(s,c|x)]\\
   &\qquad\qquad-\log (p(s)p(c))\big)\\
 & = \int_s\int_c p(s,c)\big(\log \mathbb E_x[p(s|x)p(c|x)]\\
 &\qquad\qquad-\log (\mathbb E_x[p(s|x)]\mathbb E_x[p(c|x)])\big)\\
\end{aligned}
\end{equation}
It is easy to show that minimizing $I(s,c)$ is equivalent to minimizing the following equation:
\begin{equation}
\begin{aligned}
&\mathbb E_x[p(s|x)p(c|x)]-\mathbb E_x[p(s|x)]\mathbb E_x[p(c|x)]\\
=&\int_x p(x)p(s|x)p(c|x)-\int_x p(x)p(s|x)\mathbb E_x[p(c|x)]\\
=&\int_x p(x)p(s|x)\Big(p(c|x)-\mathbb E_x[p(c|x)]\Big)\\
\end{aligned}
\end{equation}
Similarly, the above equation can be transformed to:
\begin{equation}
\int_x p(x)p(c|x)\Big(p(s|x)-\mathbb E_x[p(s|x)]\Big)
\end{equation}
It is easy to show that minimizing these two equations is equivalent to minimizing $KL(p(c|x)||p(c))$ and $KL(p(s|x)||p(s))$, which are just the regularization term of VAE.
\end{proof}

\subsection{Proof of Theorem~\ref{the:tts}}\label{tts}
\begin{proof}
According to the condition $\mathcal H(p(t_j|s_i))=MAX$, we have $p(t_j|s_i)=p(t_j), j\ne i$. Then,
\begin{equation}
p(s_j|t_i)=\frac{p(t_i|s_j)p(s_j)}{p(t_i)}=p(s_j)
\end{equation}
Consider the KL divergence between $p(t_i|s_i)$ and $p(t_i|t_j,s_i)$:
\begin{equation}
\begin{aligned}
&KL(p(t_i|s_i)||p(t_i|t_j,s_i))\\
&=-\sum_{t_i}p(t_i|s_i)\Big(\log p(t_i|t_j,s_i)-\log p(t_i|s_i)\Big)\\
\end{aligned}
\end{equation}
Then, $$\mathcal H(p(t_i|s_i))=-\sum_{t_i}p(t_i|s_i)\log p(t_i|s_i)=0$$ 
Therefore,
\begin{equation}
\begin{aligned}
&KL(p(t_i|s_i)||p(t_i|t_j,s_i))\\
&=-\sum_{t_i}\Big(p(t_i|s_i)\log p(t_i|t_j,s_i)\Big)\\
&=-\sum_{t_i}\Big(p(t_i|s_i)\log \frac{p(s_i|t_i,t_j)p(t_i)}{p(s_i)}\Big)\\
&=-\sum_{t_i}\Big(p(t_i|s_i)\log \frac{p(s_i|t_i)p(t_i)}{p(s_i)}\Big)\\
&=-\sum_{t_i}\Big(p(t_i|s_i)\log p(t_i|s_i)\Big)=0\\
\end{aligned}
\end{equation}
Therefore, $p(t_i|s_i)=p(t_i|t_j,s_i)$, so we obtain  $\mathbb E_{s_i}p(t_i|s_i)=\mathbb E_{s_i}p(t_i|t_j,s_i)$, which gives $p(t_i|t_j)=p(t_i)$.
\end{proof}
 
%
%


\subsection{Proof of Theorem~\ref{the:21}}\label{21}
\begin{theorem}\label{the:21}
If $\forall s'\in p(s_i|t_i)$, $\langle s_j, s'\rangle=0$~($j\neq i$), and intra-distinguishable condition is satisfied: $\forall p'\in \mathbb R_p$ ($\mathbb R_p$ represents the set of probability distribution vectors), $\mathcal{H}[p(t_i|s_i)] \le \mathcal{H}[p']$, then  we also have the inter-indistinguishable condition satisfied: $\forall p'\in \mathbb R_p$, $\mathcal{H}[p(t_i|s_j)] \ge \mathcal{H}[p']$.
\end{theorem}
\begin{proof}
Since $\forall s'\in p(s_i|t_i), \langle s_j, s'\rangle=0$, assume that for a specific style \type\ $t'_i$, $p(s_i|t'_i)$ has mean $\mu_i$ and covariance matrix $\Sigma_i$,  therefore any vector sampled from this distribution can be represented as $\mu_i+A_i\epsilon$, where $\epsilon$ is a random vector, $A_i^\top A_i=\Sigma_i$. 

Given the length of vector $\mu_i$ as $k$, then assume the following is a group of basis of the $k$-dim space:
\begin{equation}\label{eq:basis}
\begin{aligned}
\mu_i&+A_i\epsilon^{(1)}\\
\mu_i&+A_i\epsilon^{(2)}\\
&\ldots\\
\mu_i&+A_i\epsilon^{(k)},
\end{aligned}
\end{equation}
where $\epsilon^{(k)}$ is a one-hot vector with only the $k$-th element. Since $s_j$ is perpendicular to all of them, then  we have $k+1$ basis vectors in the $k$-dim space, which is impossible. Thus, the vectors listed in Eq.~\ref{eq:basis} are linearly related, which means that the rank of covariance matrix $rank(\Sigma_{i})\le k-1$. Then, according to Eq.~\ref{eq:gauss}, we have $p(t'_i|s_j)=0$. Finally, $\mathcal{H}[p(t_i|s_j)]$ achieves the largest value.
\end{proof}

\begin{table}[!t]
\centering
\begin{small}
\begin{tabular}{|c|c|c|c|c|}
\hline
\multicolumn{2}{|c|}{}    &\multicolumn{3}{|c|}{Tense}\\
\cline{3-5}
\multicolumn{2}{|c|}{}       & Past & Now & Future\\
\hline
\multirow{2}{*}{Sentiment}&Pos &  93516      & 285773       &20711\\
\cline{2-5}
                                     &Neg&  133388       & 244758        &21854\\
\hline
\end{tabular}
\end{small}

\medskip 
\caption{Summary statistics for the tense labels of the Amazon dataset.}
\label{tab:amazontense}
\end{table}%

\section{Case Study}

We list some of the style transferred results in Table~\ref{tab:morecase}. For sentiment, we listed the results of transfer directions ``Pos$\rightarrow$Neg" and ``Neg$\rightarrow$Pos". For tense, we cycled transfer the tense \types: ``Past$\rightarrow$Now'', ``Now$\rightarrow$Future'', and ``Future$\rightarrow$Past''. 
\begin{table*}[htp]
\begin{center}
\resizebox{\linewidth}{!}{
\begin{tabular}{p{6cm}p{6cm}p{6cm}}
\toprule[1.0pt]
\textbf{Original (Positive)} & \textbf{Vanilla Transferred (Negative)} & \textbf{VAE Transferred (Negative)}\\
\midrule[1.0pt]
the staff is awesome and i feel very comfortable leaving my babies with them ! & the staff is very rude and i will never go back here again !  & the staff is extremely rude and the customer service does n't have any sense . \\
\midrule[0.5pt]
excellent service , too ! & terrible service , too !  & horrible service , too ! \\
\midrule[0.5pt]
we would highly recommend them to all of our family and friends . & we would not recommend this place to anyone who needs any reason . & we would not recommend them to anyone who has any problems . \\
\midrule[0.5pt]
the food is excellent not to mention the staff !  & the food is not worth the money for me .  & the food is in very poor quality at a great price .\\
\midrule[0.5pt]
very thankful for this incredible place !  & very disappointed for this place .  & not a fan of this place ! \\
\midrule[1.0pt]\midrule[1.0pt]
\textbf{Original (Negative)} & \textbf{Vanilla Transferred (Positive)} & \textbf{VAE Transferred (Positive)}\\
\midrule[0.5pt]
overall service was just so-so ( except for the spa ) . & overall service was just as good as the food in vegas .  & overall service was great and we had a lot of fun . \\
\midrule[0.5pt]
the front desk person appeared not to care . & the front desk staff was friendly and helpful .  & the front desk staff was very nice and helpful .\\
\midrule[0.5pt]
everything about the room was so tired , worn out and dirty . & everything about the place was so clean and the food was very good . & everything about the place was clean and well kept as well .\\
\midrule[0.5pt]
this place is dirty and old inside . & this place is clean and tidy .  & this place is clean and comfortable . \\
\midrule[0.5pt]
this was an awful experience , i'll never stay here again .  & this experience was a great experience from the staff ! & this was definitely an excellent place to eat and enjoy !\\
\midrule[1.0pt]
\midrule[1.0pt]

\textbf{Original (Past)} & \textbf{Vanilla Transferred (Now)} & \textbf{VAE Transferred (Now)}\\
\midrule[0.5pt]
i gave this filter 5 stars because it does exactly what its suppose to . & i am sure this product would do as expected .&i have had this filter for a few months now and it 's still working great . \\
\midrule[0.5pt]
i wore this product today for a few hours today . & i wear this product today . & i wear this product for a long time today . \\
\midrule[0.5pt]
so , i searched amazon and found a real deal . & so , i 'm very happy with the purchase price . & so , i 'm very pleased with the quality of this product . \\
\midrule[1.0pt]
\midrule[1.0pt]
\textbf{Original (Now)} & \textbf{Vanilla Transferred (Future)} & \textbf{VAE Transferred (Future)}\\
\midrule[0.5pt]
this is the worst product i have ever purchased . & i will never buy this product again & i will not buy this again\\
\midrule[0.5pt]
i am not satisfied with the performance of this at all . & i am not going to buy the same brand of this . & i will never buy this again in the future \\
\midrule[0.5pt]
this barely kept things cool for four hours , with ice in it . & this will not keep things cool with ice to use at all . & It will not keep things cool with ice in it .\\
\midrule[1.0pt]
\midrule[1.0pt]
\textbf{Original (Future)} & \textbf{Vanilla Transferred (Past)} & \textbf{VAE Transferred (Past)}\\
\midrule[0.5pt]
they will most likely work out within a few days . i suggest buying this product . &they were all great as they were a gift bought from my local store  . & i bought this for my daughter and she loves it and it is great .\\
\midrule[0.5pt]
my kittens will play with just about anything . & my kittens started play this for about 2 months .&my kittens played with a lot of toys in christmas . \\
\midrule[0.5pt]
it will fit in nicely with my two larger cookers .& it was an excellent case on my cookers . & it was a little bit of good fit on my cookers . \\

\bottomrule[1.0pt]
\end{tabular}
}
\end{center}
\caption{Examples of style transfer results.}
\label{tab:morecase}
\end{table*}%

\begin{table*}[!t]
\begin{center}
\resizebox{\linewidth}{!}{
\begin{tabular}{p{6cm}p{6cm}p{6cm}}
\toprule[1.0pt]
\textbf{Original (Pos, Now)} & \textbf{Vanilla Transferred (Neg, Past)} & \textbf{VAE Transferred (Neg, Past)}\\
\midrule[0.5pt]
this book is excellent for us self taught decorators. & this book was not very well for decorators  & this book was extremely nonsense and totally not a good choice for us . \\
\midrule[0.5pt]
i am thoroughly enjoying being able to learn at my own pace . & i was not happy on learning by myself . &i was not happy at all when learning this book . \\
\midrule[0.5pt]
the book is very easy to follow with lots of step by step pictures . & the book was hard for  cookers . & the book was difficult to read , not a good choice for cookers . \\
\midrule[1.0pt]
\midrule[1.0pt]
\textbf{Original (Pos, Now)} & \textbf{Vanilla Transferred (Neg, Future)} & \textbf{VAE Transferred (Neg, Future)}\\
\midrule[1.0pt]
i'm glad i purchased this item on amazon.com .	& i will regret to purchase this  on amazon.com  .&  i will regret to purchase such kind of product  on amazon.com .\\
\midrule[1.0pt]
we've had a lot of fun and good eating from this machine.  & we will never eat  from this machine again . & we will regret to eat food from this machine tomorrow . \\
\midrule[1.0pt]\midrule[1.0pt]
\textbf{Original (Neg, Future)} & \textbf{Vanilla Transferred (Pos, Past)} & \textbf{VAE Transferred (Pos, Past)}\\
\midrule[1.0pt]
i'm going to try again but seriously thinking of sending it back. & i tried this and decide to keep it . & i had tried this many times and think it was a good product . \\
\midrule[1.0pt]
i will discard the pan and either buy a good quality pressure cooker or do without one.	 & the pan was in good quality . & i kept the pan to make it a good cooker .\\
\midrule[1.0pt]
i will never recommend this to any one.	& i recommended this to my friend . & i had recommended this to all my friends .\\
\bottomrule[1.0pt]
\end{tabular}
}
\end{center}
\caption{The examples of transferring multiple \genres\ simultaneously.}
\label{tab:morecasemg}
\end{table*}%

\section{More Data Preprocessing Details}
 Yelp Service Reviews  is already split into training, validation, and test sets, containing 444,101, 63,483, and 126,670 labeled reviews, respectively. 
Similarly, Amazon Product Reviews contains 559142, 2000, and 2000 labeled reviews as training, validation, and test sets, respectively. 
The statistical information of tense labeling is shown in Table~\ref{tab:amazontense}.
\section{Implementation details}

The encoder and decoder are set as 2-layer LSTM RNNs with input dimension of 100, the hidden size is 150, and max sample
length is 15. Balancing parameters are set to $\lambda_a=1.0, \lambda_c=0.5, \lambda_s=0.8, \lambda_\text{KL}=0.01, \lambda_\text{sc}=0.1$ and $\lambda_m=0.4$. Each hyperparameter are chosen by grid search. For $\lambda_\text{KL}$, the search scope is $[0.0, 0.1]$, step size is $0.01$. For other hyperparameters, the search scope is $[0.0, 1.0]$, step size is $0.1$. The search scope and  step size are chosen by experience.

\section{Human Evaluation Question Marks}
\subsection{Transfer Accuracy (TA)}
\subsubsection{Sentiment}
Q: Do you think the given sentence belongs to positive sentiment or negative sentiment?

\begin{itemize}
\item A: Positive. 
\item B: Negative.
\end{itemize}

\subsubsection{Tense}

Q: Do you think the given sentence belongs to past, now, or future?

\begin{itemize}
\item A: Past. 
\item B: Now.
\item C: Future.
\end{itemize}

\subsection{Content Preservation (CP)}

Q: Do you think the generated sentence has the same content with the original sentence, although the sentiment/tense is different?

Please choose a score according to the following description. Note that the score is not necessary to be integer, you can give scores like $3.2$ or $4.9$ by your feeling.
\begin{itemize}
\item 5: Exactly. The contents are exactly the same.
\item 4: Highly identical. Most of the content are identical.
\item 3: Half. Half of the content is identical.
\item 2: Almost Not the same.
\item 1: Totally different.
\end{itemize}

\subsection{ Language Quality (LQ)}

Q: How fluent do you think the generated text is? Give a score based on your feeling.

Please choose a score according to the following description. Note that the score is not necessary to be integer, you can give scores like $3.2$ or $4.9$ by your feeling.
\begin{itemize}
\item 5: Very fluent. 
\item 4: Highly fluent. 
\item 3: Partial fluent. 
\item 2: Very unfluent. 
\item 1: Nonsense.
\end{itemize}

\end{document}